\documentclass[10pt,twocolumn,letterpaper]{article}
\newcommand{\RGB}{\text{RGB}}
\newcommand{\LAB}{\text{LAB}}
\newcommand{\XYZ}{\text{XYZ}}
\usepackage{cvpr}
\usepackage{subcaption}
\usepackage{caption}
\usepackage{times}
\usepackage{epsfig}
\usepackage{graphicx}
\usepackage{amsmath}
\usepackage{amssymb}
\usepackage{float}

\usepackage[pagebackref=true,breaklinks=true,colorlinks,bookmarks=false]{hyperref}

 \cvprfinalcopy 

\def\cvprPaperID{****} 

\ifcvprfinal\fi

\title{Color and Edge-Aware Adversarial Image Perturbations}
\author{
Robert Bassett\\
Naval Postgraduate School\\
{\tt\small robert.bassett@nps.edu}
\and
Mitchell Graves\\
US Naval Academy\\
{\tt\small mgraves@usna.edu}
\and
Patrick Reilly\\
Naval Postgraduate School\\
{\tt\small patrick.reilly@nps.edu}
}

\begin{document}
\twocolumn[{%
\renewcommand\twocolumn[1][]{#1}%
\maketitle
\begin{center}
    \centering
  \minipage{.33\textwidth}
    \centering
    \includegraphics[width=\textwidth]{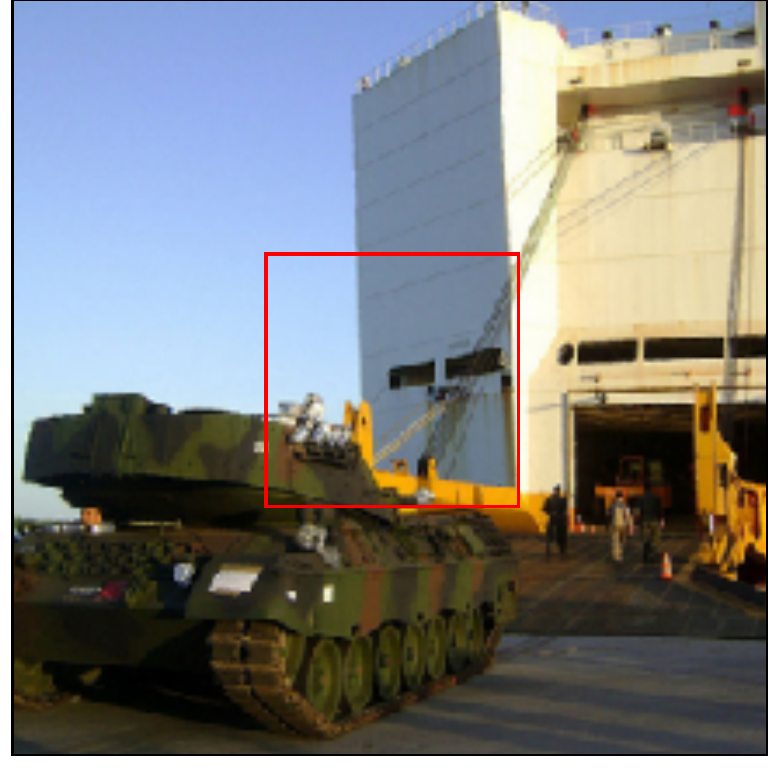}
  \endminipage
\minipage{.33\textwidth}
    \includegraphics[width=\textwidth]{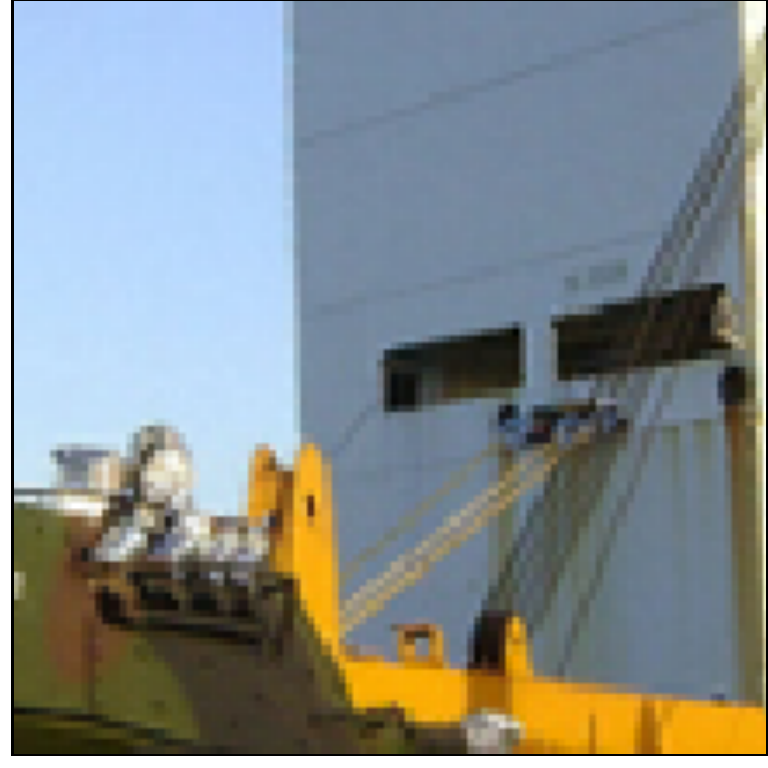}
  \endminipage
\minipage{.33\textwidth}
    \includegraphics[width=\textwidth]{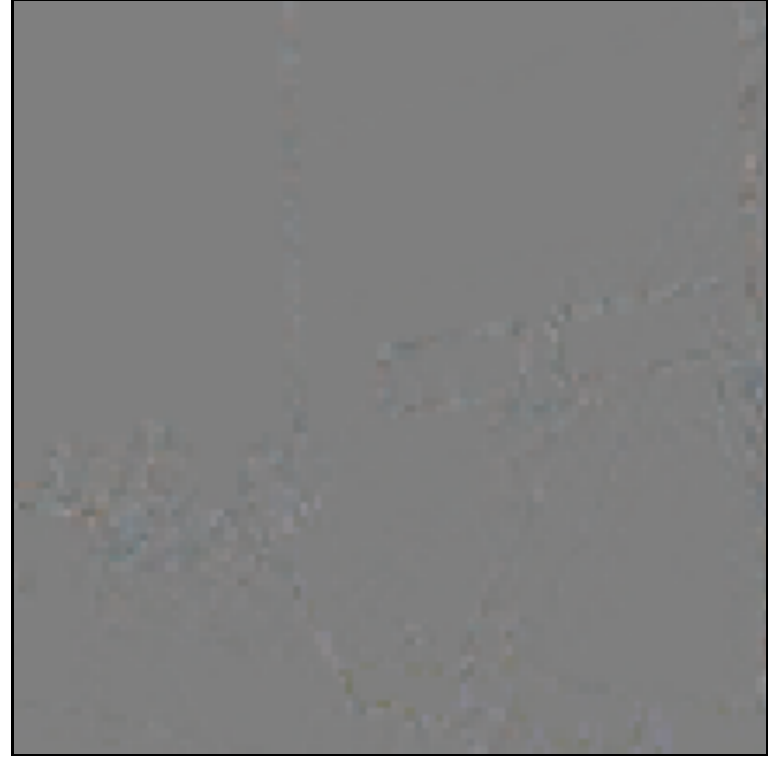}
\endminipage
  \captionof{figure}{A Color-and-Edge-Aware Perturbation. Left: an image of a tank which has been adversarially perturbed through a targeted misclassification so that it is classified as an amphibious vehicle with $98.7\%$ confidence. Center: A region of the left image. Right: The image perturbation, which has been constructed to model human perception of texture and color. Image source: ILSVRC 2012 \cite{imagenet}. Classifier: Inception v3 \cite{inceptionv3}.}
\label{introfig}
\end{center}%
}]

%
\begin{abstract}
  Adversarial perturbation of images, in which a source image is deliberately modified with the intent of causing a classifier to misclassify the image, provides important insight into the robustness of image classifiers. In this work we develop two new methods for constructing adversarial perturbations, both of which are motivated by minimizing human ability to detect changes between the perturbed and source image. The first of these, the \emph{Edge-Aware} method, reduces the magnitude of perturbations permitted in smooth regions of an image where changes are more easily detected. Our second method, the \emph{Color-Aware} method, performs the perturbation in a color space which accurately captures human ability to distinguish differences in colors, thus reducing the perceived change. The Color-Aware and Edge-Aware methods can also be implemented simultaneously, resulting in image perturbations which account for both human color perception and sensitivity to changes in homogeneous regions. Because Edge-Aware and Color-Aware modifications exist for many image perturbations techniques, we also focus on computation to demonstrate their potential for use within more complex perturbation schemes. We empirically demonstrate that the Color-Aware and Edge-Aware perturbations we consider effectively cause misclassification, are less distinguishable to human perception, and are as easy to compute as the most efficient image perturbation techniques. Code and demo available at 
\href{https://github.com/rbassett3/Color-and-Edge-Aware-Perturbations}{https://github.com/rbassett3/Color-and-Edge-Aware-Perturbations}.
\end{abstract}

%
%

\section{Introduction}

Adversarial perturbations have shown that state-of-the-art techniques for image classification are inherently unstable, because minute changes to an image can result in dramatic changes in the predicted class of the image. Many techniques have been introduced to generate adversarial perturbations, but a common theme is a formulation which encourages substantial change to the output of the classifier while restricting to only small changes of the image. In these formulations, metrics for quantifying change to the image are often mathematically instead of perceptually motivated.

We address this problem by proposing two new techniques for generating adversarial perturbations. The first, our Edge-Aware method, is motivated by human ability to detect minor modifications against a smooth background. It uses a texture filter, such as a Sobel or Gabor filter, to limit perturbations in smooth regions. The result is that the Edge-Aware method constructs perturbations which preserve smoothly textured regions in the image. Though this limitation might be expected to make misclassification more difficult to achieve, we find that misclassification can still be caused relatively easily, even when generating perturbations which target a certain class for the perturbed image.

Our second contribution is the Color-Aware method for generating image perturbations. While the Edge-Aware method reduces detection by considering how a pixel differs from its neighbors, the Color-Aware method focuses on the pixel value itself. It is well-known that for RGB representations of a pixel, metrics like $\ell_2$ or $\ell_{\infty}$ do not accurately capture human ability to perceive color difference. Therefore perturbations which are small with respect to these metrics may still be easily detected by an individual comparing the perturbed and source images. To overcome this issue we convert the image to a color space which \emph{does} capture human perception in color difference, and construct the perturbation in this space. One concern with this approach is computational, because conversion from RGB to color spaces which accurately model human perception can involve complicated transformations. We mitigate this concern by using the CIE L*a*b* (CIELAB) color space, in which the $\ell_2$ distance between pixels captures perceived color distance. The tractability of this constraint, and the fact that we construct the perturbation directly in CIELAB color space, allows us to construct Color-Aware perturbations with minimal computational overhead.

The Color-Aware and Edge-Aware methods can be applied simultaneously to generate a Color-and-Edge-Aware perturbation, an example of which appears in Figure \ref{introfig}. Color-and-Edge-Aware perturbations reduce human ability to distinguish between the source and modified images by constraining both texture and color discrepancies. In the remainder of this paper, we demonstrate the effectiveness of Color and Edge-Aware perturbations. We show that they are among the most computationally efficient methods for generating adversarial perturbations, while still effectively causing misclassification of the perturbed image. We also provide numerical experiments which confirm that Color and Edge-Aware perturbations are effective in applications where additional human scrutiny is expected.

Before proceeding we establish some notation. We assume that images have been scaled to take values in $[0,1]$. Let $w$, $h$, and $c$ be fixed positive integers which give the width, height, and number of color channels, respectively, of the images considered. Denote $[0,1]^{h \times w \times c}$, the set of valid images, by $\mathcal{X}$. Let $\mathcal{C} := \{1, ..., C\}$ be a set of possible image classes. An image classification algorithm is a function $F: \mathcal{X} \to \Delta^{C}$, where $\Delta^{C}$ denotes the $C$-dimensional probability simplex. It is common for an image classifier to convert logits to probabilities using the softmax function, in which case $F$ can written $F = \text{softmax} \circ Z$, for some function $Z: \mathcal{X} \to \mathbb{R}^{C}$. Throughout, we will denote a source image by $x$, a perturbation by $\delta$, and a perturbed image $x + \delta$ by $x'$. We denote by $\|\cdot \|_{p, c} : [0,1]^{h \times w \times c} \to \mathbb{R}^{h \times w}$ an $\ell_{p}$ norm applied across the color dimension of an image. Otherwise, $\|\cdot \|_{p} \to \mathbb{R}$ denotes the entrywise $p$-norm of a tensor. Lastly, we use $\langle \langle \cdot, \cdot \rangle \rangle$ to denote the Frobenius inner product.

\section{Related Work}

Neural networks are state-of-the-art tools for image classification, and have been used in a variety of application areas including computer vision \cite{alexnet}, \cite{zhao2019}, natural language processing \cite{goldberg}, and Markov decision processes \cite{sutton}. Despite their success, Szegedy et al. \cite{szegedy} first noticed that neural networks are vulnerable to adversarial perturbations, in which noise is added to an input in order to cause misclassification without substantially changing the input. Adversarial perturbations are especially interesting in the context of image classification, in part because of the high-level of performance that artificial neural networks enjoy in that domain. In their paper, Szegedy et al. solved for a perturbed image $x'$ of an input image $x$ by solving the following optimization problem using bound-constrained L-BFGS.
$$\min_{x'\in \mathbb{R}^{h \times w \times c}} \; L\left(F(x'), l \right) + \frac{\alpha}{2} \left\| x-x' \right\|_{2}^{2}$$
$$\text{subject to } 0 \leq x' \leq 1.$$
The function $L: \Delta^{C} \times \mathcal{C} \to \mathbb{R}$, and label $l \in \mathcal{C}$ take two forms. In the untargeted setting, where any misclassification is acceptable, $L$ is taken to be the \emph{negative} cross entropy loss and $l$ the true label of the image. In the targeted setting, $L$ is taken to be the cross entropy loss and $l$ the targeted label for the image. Large values of the parameter $\alpha$ encourage the perturbation to be close to the source image; this value must be chosen separately.

A few features of the Szegedy et al. method are worth emphasizing. First, the optimization method is quasi-newton, and hence requires only first-order information about the objective function. This is important because modern software for neural networks emphasizes efficient gradient computation. Second order information, on the other hand, would be extremely burdensome to compute and cannot be assumed. Another important feature of the Szegedy et al. method is that projection onto the constraint set is easy, so that it can be computed quickly as part of iterative first-order methods. Lastly, we note that, despite its simplicity, the L-BFGS method requires extra memory to store Hessian approximations, and can also require many function evaluations to compute the step length (often via backtracking).

In contrast to the L-BFGS method of Szegedy et al., the Fast Gradient Sign method (FGSM), prioritizes perturbations which can be easily computed \cite{FGSM}. The FGSM method proposes the following optimization problem
$$\min_{\delta \in \mathbb{R}^{h \times w \times c}} \; L\left(F(x + \delta), l \right)$$
$$\text{subject to } \|\delta\|_{\infty} < \alpha.$$
To reduce the computational burden associated with solving this problem, the authors instead optimize a linear approximation of the objective by taking the gradient of the input with respect to $\delta$.
$$\min_{\delta \in \mathbb{R}^{h \times w \times c}} \; \langle \langle \nabla_{x} \left(L\left(F(x), l \right)\right) , \delta \rangle \rangle $$
$$\text{subject to } \|\delta\|_{\infty} < \alpha.$$
This has the closed-form solution 
$$x' = x + \alpha \, \text{sign}\left(\nabla_{x} \left(L\left(F(x), l \right)\right)\right).$$
In the event that $x' \not \in [0,1]^{h \times w \times c}$, it can easily be projected onto this set, making the result a valid image. FGSM can also be used as an iterative method, where the perturbed image $x'$ from one iteration is used as the input image $x$ in the next iteration.

The FGSM's simplicity is the key to its success. Though the linear approximation of FGSM is simpler than the quasi-newton method of Szegedy et al, FGSM requires only the elementwise sign of the gradient yet has been shown to still generate effective perturbations. 

There have been many other methods proposed to generate adversarial perturbations. Two of the most celebrated are the Carlini-Wagner $\ell_{2}$ perturbation \cite{carlini-wagner} and DeepFool \cite{DeepFool}. The Carlini-Wagner approach is designed to achieve very precise misclassification, and it includes a tuning parameter which specifies the misclassification confidence. In this sense, the Carlini-Wagner approach is well-suited to answer the question: ``What is the minimal perturbation required to move this image onto the decision boundary between classes?'' Though it effectively generates adversarial perturbations, the Carlini-Wagner method is complex relative to other methods for adversarially perturbing images, requiring multiple starting points and at least three additional univariate parameters depending on the descent method used. Furthermore, our proposed methods can be easily incorporated in a Carlini-Wagner type perturbation by composing the classifier with a color space transformation and appropriately weighting the $\ell_{2}$ norm. Because of their complex formulation and relatively high computational overhead, Carlini-Wagner perturbations have different motivations than Color-Aware and Edge-Aware perturbations as considered in this paper, in which we seek to \emph{efficiently} and effectively create image perturbations which are undetectable by human observers. Doing so will accomplish two goals. The first is practical; perturbations which are easier to be compute can be more readily applied. Our second goal is theoretical, in that we seek the simplest technique that accomplishes the task of constructing effective and imperceivable image perturbations. This tractability gives our contributions additional utility because they can be used within more complex perturbation schemes.

The motivation for DeepFool better aligns with the goals of this paper because of its emphasis on lightweight construction of perturbations. The DeepFool algorithm proceeds iteratively by stepping towards the decision boundary of the classifier. In order to make these iterates tractable, DeepFool linearly approximates the decision boundary at each iteration. DeepFool prioritizes efficient computation, and in this way is a compromise between Carlini-Wagner and FGSM. One drawback of DeepFool is that it only accommodates untargeted perturbations. Like DeepFool, we are motivated by efficient computation of perturbations, but our method will accommodate targeted perturbations. We also note that DeepFool, like Carlini-Wagner and many other perturbation methods, can be easily modified to include both our Edge-Aware and Color-Aware ideas by changing the norm it uses internally.

\section{Proposed Approaches}

We begin by describing our Color-Aware perturbation. The primary motivation for developing our Color-Aware perturbation is that the distance between colors, when represented as vectors in RGB space, does not correspond to the perceived difference from the perspective of a human observer. There have been many efforts to devise color systems which respect human perception, beginning with the Munsell Color System \cite{Munsell} in 1905. Since then, the International Commission on Illumination (with acronym CIE in French) developed a sequence of color spaces and color distances which quantify perceived color difference. The first of these was in 1931 with the CIEXYZ color space. This space was improved in 1976 with the addition of the CIELAB and CIELUV color spaces, which better model perceived color difference. The CIELAB and CIELUV spaces perform similarly with respect to their accuracy in perceived color difference \cite{MorleyData}, and we opt to use CIELAB. We note that in the CIELAB representation of colors, perceived distance between colors is measured using the Euclidean distance between them. 

\begin{figure}[h]
  \minipage{.16\textwidth}
    \includegraphics[width=\textwidth]{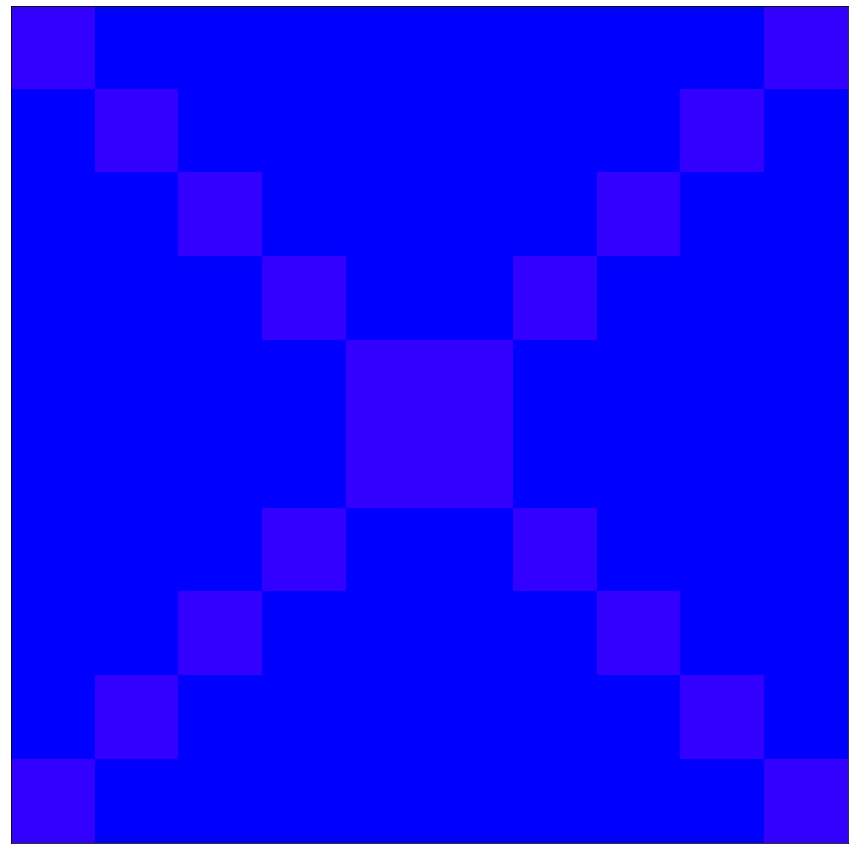}
  \endminipage
\minipage{.16\textwidth}
    \includegraphics[width=\textwidth]{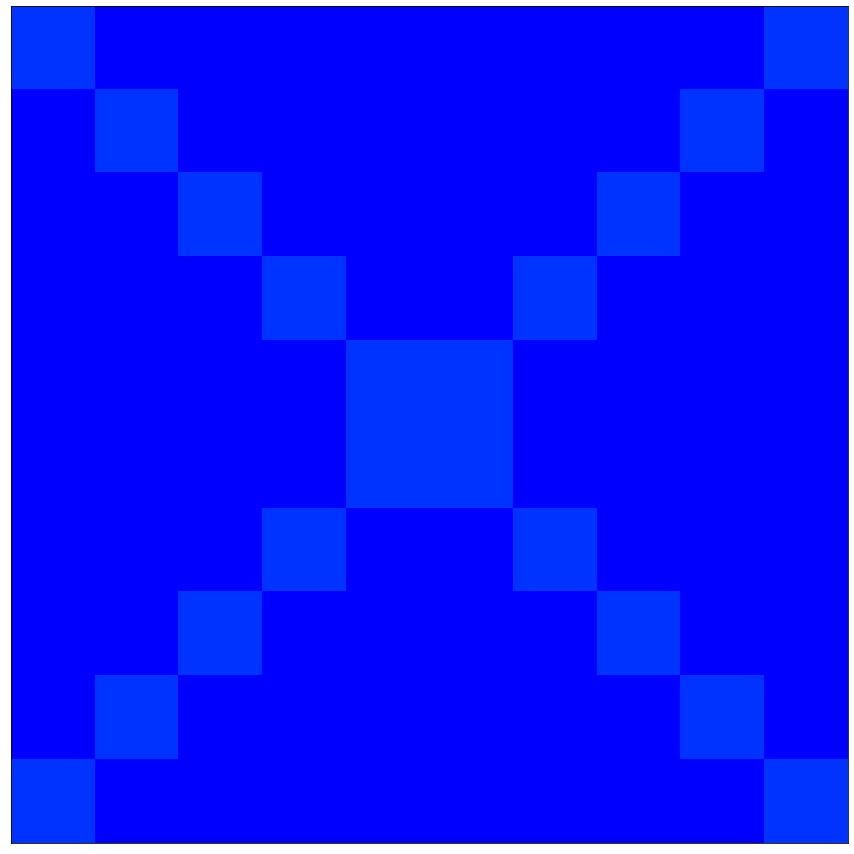}
  \endminipage
\minipage{.16\textwidth}
    \includegraphics[width=\textwidth]{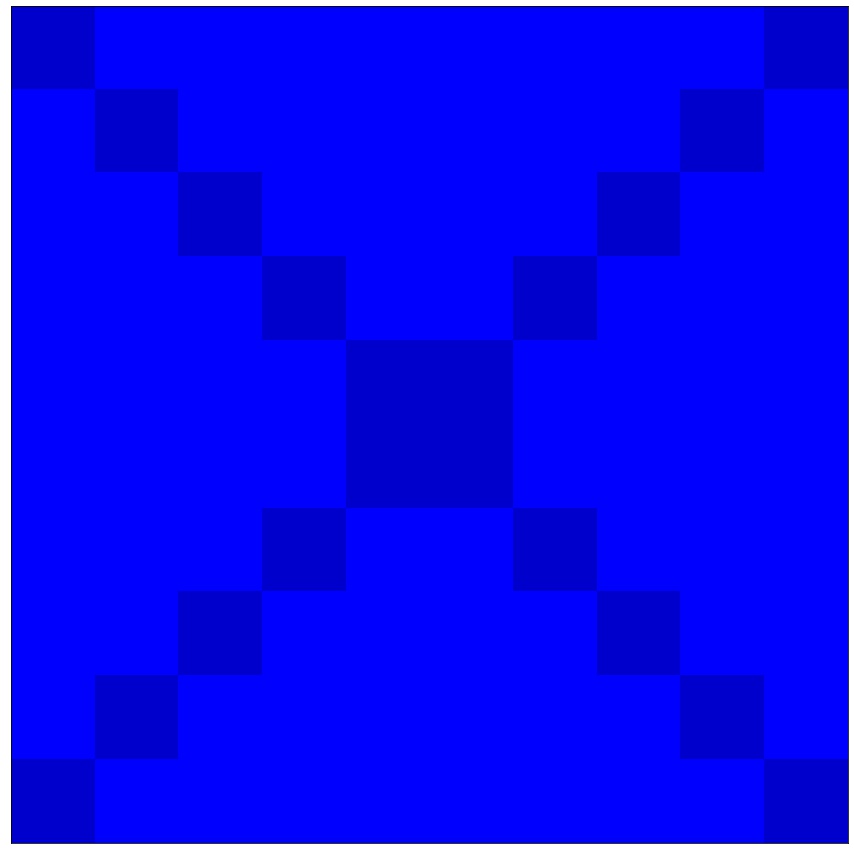}
  \endminipage
  \caption{An illustration of the CIELAB color distance. Each `x' is formed by changing the solid blue background, $(0,0,1)$ in RGB, by $\pm .2$ in a single color plane. Though equidistant in RGB, some of the 'x's are more easily distinguished from the background. The CIELAB distance, on the other hand, accurately captures perceived color difference. Left: CIELAB distance 3.04. Center: CIELAB distance 17.23. Right: CIELAB distance 76.94.}
  \label{cielabfig}
\end{figure}

Conversion from RGB to CIELAB requires an intermediate conversion to CIEXYZ, which is a linear transformation.
\begin{align}
  {\begin{bmatrix}X\\Y\\Z\end{bmatrix}} = \mathbf{A} {\begin{bmatrix}R\\G\\B\end{bmatrix}}
\end{align}
The (invertible) matrix $\mathbf{A} \in \mathbb{R}^{3\times 3}$ is specified in the CIEXYZ standard \cite{CIEXYZ}. Conversion from XYZ to LAB space is nonlinear.
\begin{align}
  L^{*} &= 116 f\left(\frac{Y}{Y_{n}} \right) - 16 \\
  a^{*} &= 500 \left( f\left(\frac{X}{X_{n}}\right) - f\left(\frac{Y}{Y_{n}}\right)\right) \\
  b^{*} &= 200 \left( f\left(\frac{Y}{Y_{n}}\right) - f\left(\frac{Z}{Z_{n}}\right)\right)
\end{align}
where $\delta = \frac{6}{29}$ and
$$f(t) = \begin{cases} \sqrt[3]{t}  & \text{ if } t > \delta^{3}\\
  \frac{t}{3 \delta^{2}} + \frac{4}{29} & \text{ otherwise}.\end{cases}$$
The constants $X_n$, $Y_n$ and $Z_n$ depend on the illumination standard used, but are commonly taken as $X_{n} := 95.0489$, $Y_n := 100$, and $Z_n := 108.8840$ \cite{IlluminantD65}.
Denote by $C_{\RGB \to \XYZ}$ the conversion function from RGB to CIEXYZ, with appropriate notational extensions to other color spaces. When necessary, we will clarify the space in which a source image $x$, perturbed image $x'$, or perturbation $\delta$ reside with an appropriate superscript. We have
$$C_{\RGB \to \LAB} = C_{\XYZ \to \LAB} \circ C_{\RGB \to \XYZ}.$$
The conversion functions $C_{\XYZ \to LAB}$ and $C_{\RGB \to \XYZ}$ are invertible, so $C_{\LAB \to \RGB}$ is defined as the appropriate composition of inverses. We note that all conversion functions are continuous and piecewise differentiable.

CIELAB was further extended to color difference formulas CIEDE94 and CIEDE2000 in the corresponding years. One of the few other works to apply perceptual color-spaces to adversarial perturbations is \cite{Zhao}, which uses a Carlini-Wagner approach to compute adversarial perturbations where the difference between the perturbed and source image was quantified using CIEDE2000. That work shares our Color-Aware motivation of constructing perturbations which account for human color perception, but our emphasis on efficient computation prompts us to use CIELAB instead, thus avoiding the complexity of a Carlini-Wagner formulation. Though the CIEDE94 and CIEDE2000 difference formulas were intended to correct some imprecisions in using CIELAB to measure perceived color difference, they require complicated nonconvex manipulations of the LAB coordinates and are not given by the $\ell_{p}$ distance in some color space. This increases the computational burden required to compute adversarial perturbations and motivates our use of CIELAB. To our knowledge there have been only two other efforts using perceptual color spaces for image perturbations \cite{Athalye, Laidlaw}. Like our work, these authors use perceived color distances, but their formulations differs critically from ours because both posit an intractable constraints which require instead solving a penalized approximation to the constrained problem. Our formulation only uses tractable constraints, mirroring the simplicity of FGSM in both its constraint set and the closed-form solution of its linear approximation.

We propose to adversarially perturb an image $x^{\RGB}$ as follows. Convert $x^{\LAB} := C_{\RGB \to \LAB}(x^{\RGB})$ and solve the following.
\begin{align} \label{Color-Aware}
  \min_{\delta} L\left( F \circ C_{\LAB \to \RGB}\left( x^{\LAB} + \delta \right), l \right)\\
  \text{subject to } \|\delta\|_{2, c} \leq \alpha. \nonumber
\end{align}

As in FGSM, we linearly approximate the objective function to yield the closed-form solution below.
\begin{equation} \label{ca-pert}
x'^{\; \LAB} = x^{\LAB} + \alpha \, \frac{\nabla_{x} L\left( F \circ C_{\LAB \to \RGB}\left( x^{\LAB} \right), l \right)}{\|\nabla_{x} L\left( F \circ C_{\LAB \to \RGB}\left( x^{\LAB} \right), l \right)\|_{2, c}}
\end{equation}
We note that division in \eqref{ca-pert} is pointwise and broadcast across the color dimension so that the dimensions are compatible. Finally, the perturbed image can be converted to RGB, $x'^{\; \RGB}  = C_{\LAB \to \RGB}\left(x'^{\; \LAB}\right)$.

We remark that because $\delta$ is a perturbation in CIELAB space, the constraint in \eqref{Color-Aware} represents perceived color difference. Also, the classifier $F$ is assumed to require RGB inputs, though we note that there is work attempting to train models directly on CIELAB representations of images \cite{Diaz}.

\begin{figure*}[ht]
  \minipage{.33\textwidth}
    \includegraphics[width=\textwidth]{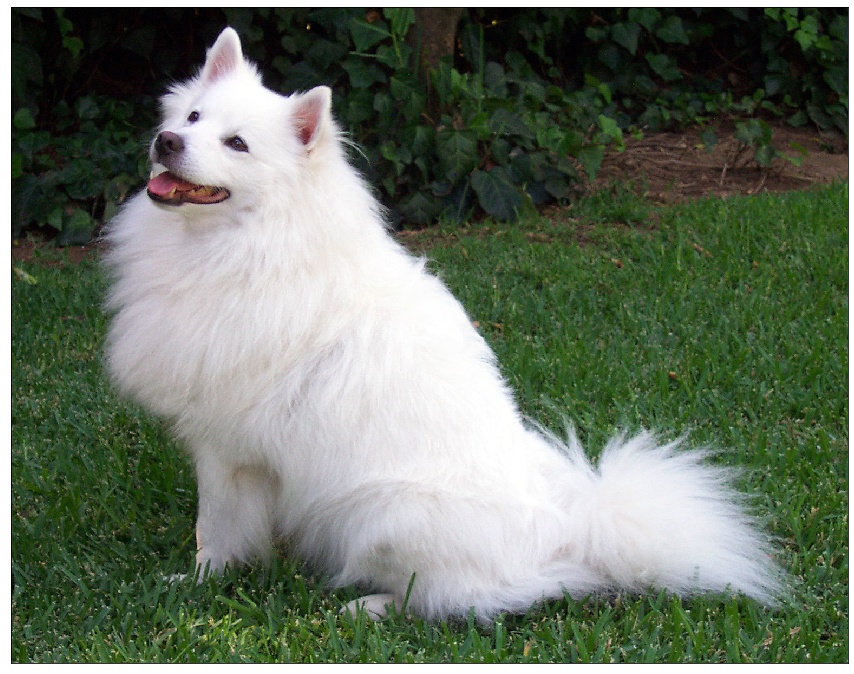}
  \endminipage%
\minipage{.33\textwidth}
    \includegraphics[width=\textwidth]{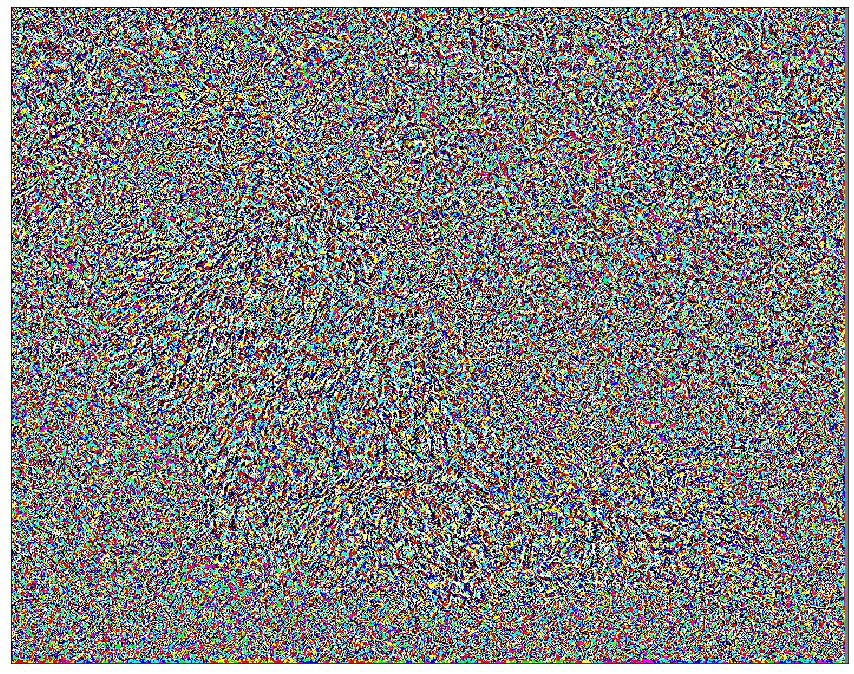}
  \endminipage%
\minipage{.33\textwidth}
    \includegraphics[width=\textwidth]{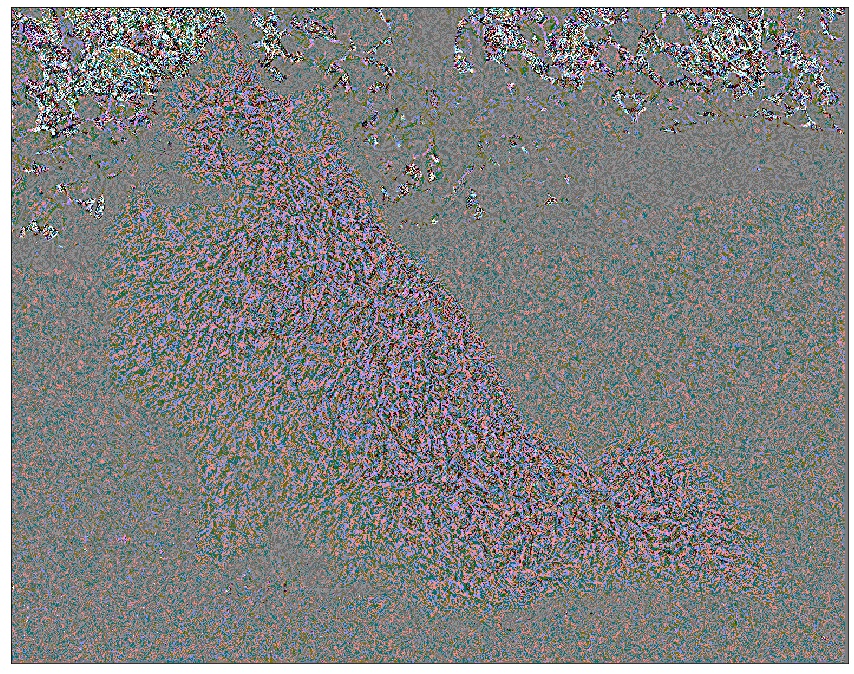}
  \endminipage
  \caption{FGSM and Color-Aware perturbations. Left: a source image to be adversarially perturbed. Center: The FGSM perturbation, scaled to $[0,1]$. Right: The Color-Aware perturbation, scaled to $[0,1]$. To allow visual comparison with FGSM, which rounds gradient values to the nearest vertex in the $\ell_{\infty}$ unit ball, we have rounded values less than $.4$ and greater than $.6$ to $0$ and $1$, respectively. Note that in the Color-Aware perturbation different objects (dog, grass, or dark background) have perturbations in different directions.}
  \label{dogfig}
\end{figure*}

Next we describe the Edge-Aware perturbation method. Let $W: \mathcal{X} \to [0,1]^{h \times w}$ denote a pixel-wise edge detector, such as the Sobel or Gabor filter, where a value near one means an edge is detected. We construct Edge-Aware perturbations by weighting pixels in the perturbation constraint by their edge weights, thus reducing the magnitude of perturbation permitted in smooth regions. This can be applied to the FGSM directly, but we will introduce it in the context of our Color-Aware perturbation method. Letting $w = W(x^{\RGB})$, we propose solving the following
\begin{align} \label{Color-and-E-Aware}
  \min_{\delta} L\left( F \circ C_{\LAB \to \RGB}\left( x^{\LAB} + \delta \right), l \right)\\
  \text{subject to } \|\delta\|_{2, c} \leq \alpha w. \nonumber
\end{align}
Again, by linearly approximating the objective function we arrive at the closed-form solution
\begin{equation}\label{main_pert}
x'^{\; \LAB} = x^{\LAB} + \alpha \,  w \, \frac{\nabla_{x} L\left( F \circ C_{\LAB \to \RGB}\left( x \right), l \right)}{\|\nabla_{x} L\left( F \circ C_{\LAB \to \RGB}\left( x \right), l \right)\|_{2, c}}
\end{equation}
$$x'^{\; \RGB} = C_{\LAB \to \RGB} \left( x'^{\; \LAB} \right).$$
As in equation \eqref{ca-pert}, the multiplication and division in \eqref{main_pert} is pointwise and broadcast across the color dimension. We also note that in equations \eqref{ca-pert} and \eqref{main_pert} it is possible that the 2-norm at a pixel is zero, in which case we do not make any perturbation at the pixel. 

\section{Numerical Experiments}

In this section we empirically evaluate the performance of our Color and Edge-Aware perturbations in two sets of experiments. The first of these compares the efficacy of our contributions using the Inception v3 classifier \cite{inceptionv3} and the ILSVRC 2012 validation set \cite{imagenet} as the classifier to disrupt and the images to perturb. For the second set of experiments we perturb images of faces--an application in which the human eye is especially capable at detecting inconsistencies--to manipulate DeepFake images with the purpose of evading a classifier built for their detection. Both applications confirm that Color and Edge-Aware perturbations are less perceptible than its competitors, while at the same time generating effective and computationally tractable perturbations.


\subsection{Imagenet \& Inception v3}


We begin by comparing perturbations computed using our Color-Aware method with those of FGSM in order to show the qualitative difference between perturbation directions. For a small value of $\alpha$, we compute $x' - x$, the difference between a perturbed and source image, where the perturbed image $x'$ is constructed through both FGSM and our Color-Aware method. Figure \ref{dogfig} gives a comparison of the perturbations, where the Color-Aware version is rounded to the nearest vertex of the RGB cube to make it visually comparable with the FGSM perturbation.

We see that our Color-Aware method identifies color trends in this perturbation that FGSM does not.  In the region containing the grass, red occurs in much lower quantities than the green and blue colors. Similar to our example in figure \ref{cielabfig}, perturbing the nondominant color planes results in large RGB change with small perceived color difference. The most prevalent colors in this region are teal $(0,.5,.5)$ and coral $(1,.5,.5)$ which represent changes in the red color plane because of the $[0,1]$ scaling. Less prevalent but still visible are lavender $(.5, .5, 1)$ and olive $(.5,.5, 0)$, which represent perturbations in the blue color plane, the other non-dominant color in the region. For the region containing the dog, the white of the dog has a fairly even distribution of RGB colors, so there is more variety in the perturbations than in the grassy region. In FGSM, however, color trends are impossible to distinguish. The region containing the dog can be distinguished upon close inspection based on the perturbation's texture, but not the distribution of its colors.

\begin{figure}[h]
\begin{center}
  \begin{subfigure}{.25\textwidth}
    \includegraphics[width=\textwidth]{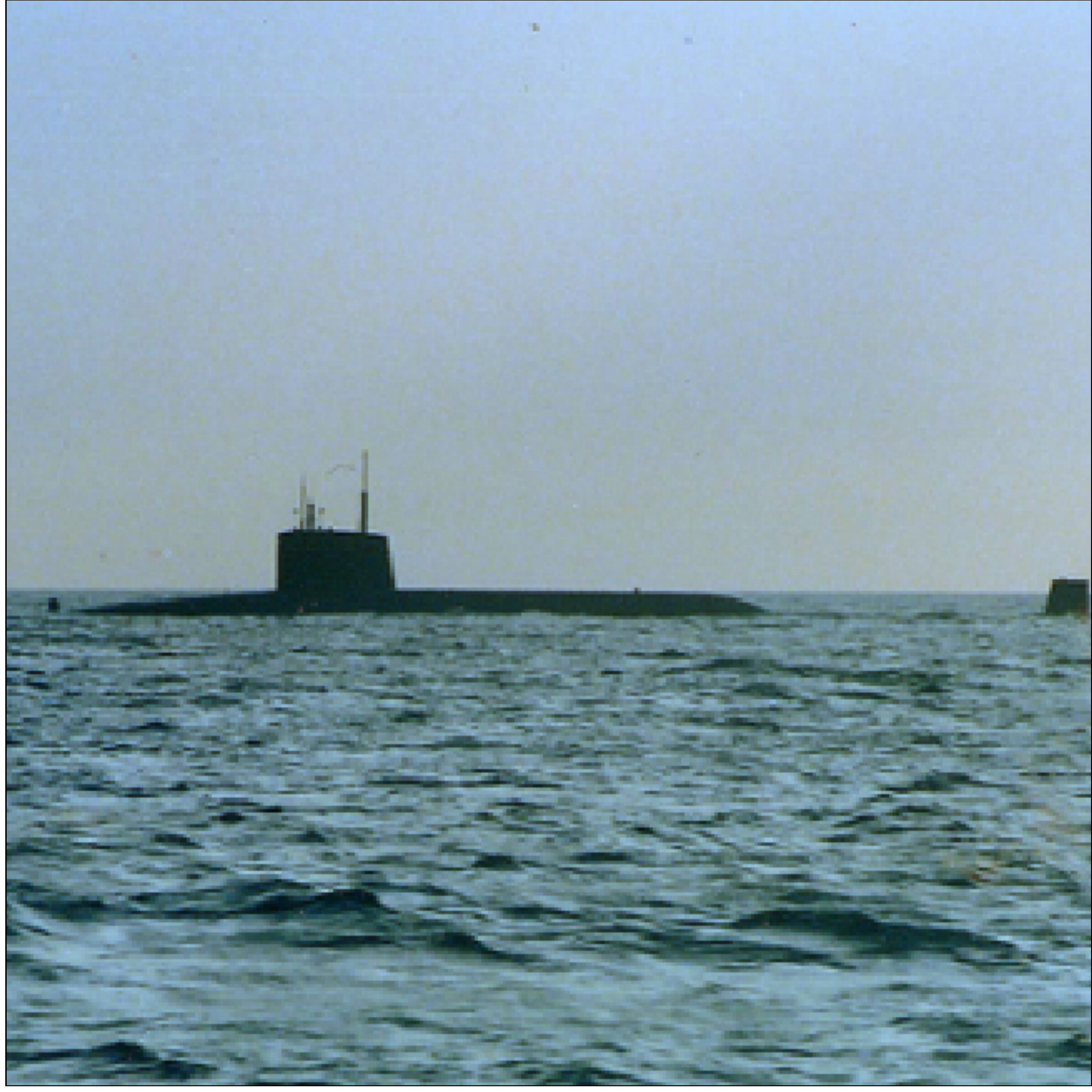}
\caption{Original image}
  \end{subfigure}%
\end{center}
  \begin{subfigure}{.25\textwidth}
    \includegraphics[width=\textwidth]{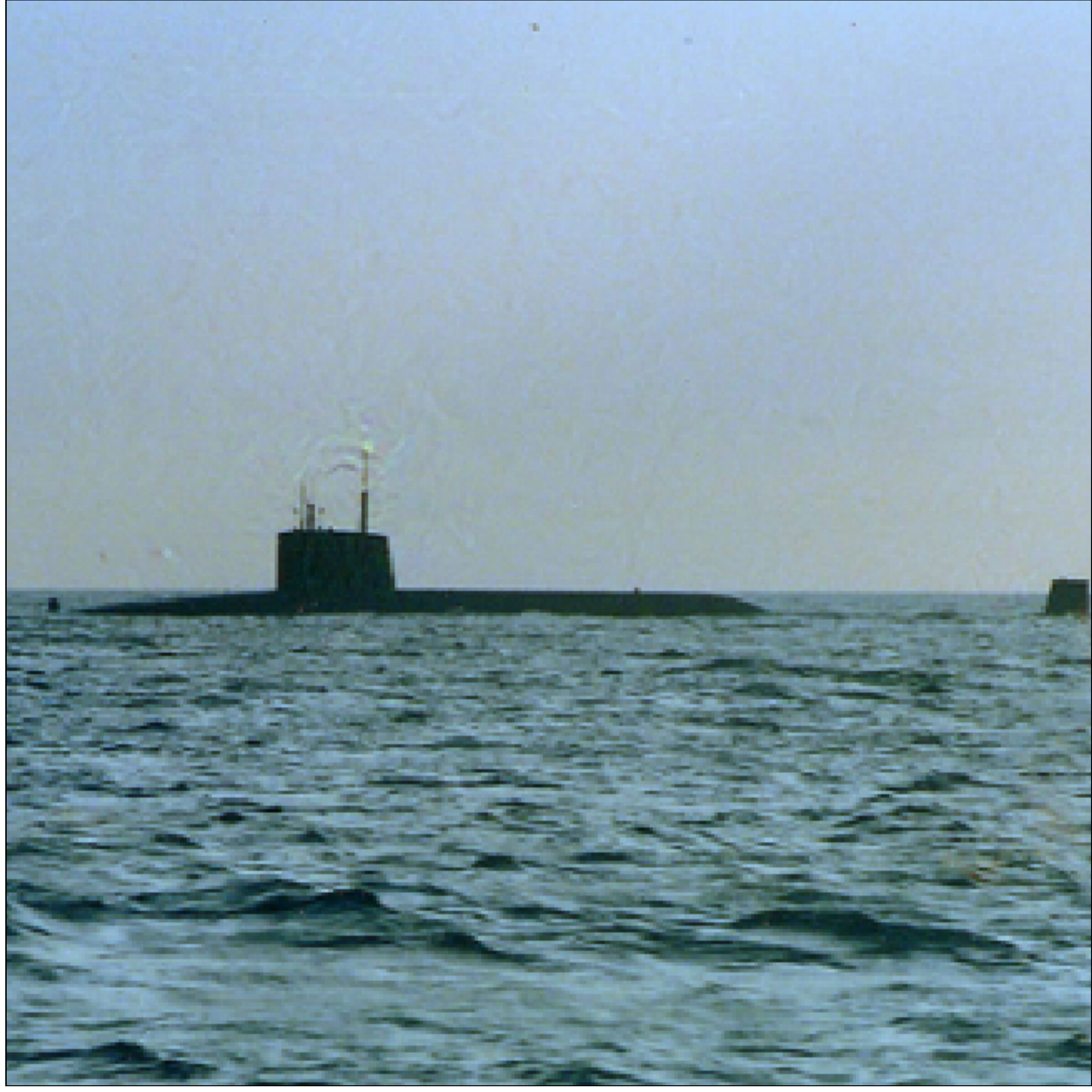}
\caption{L-BFGS}
  \end{subfigure}%
  \begin{subfigure}{.25\textwidth}
    \includegraphics[width=\textwidth]{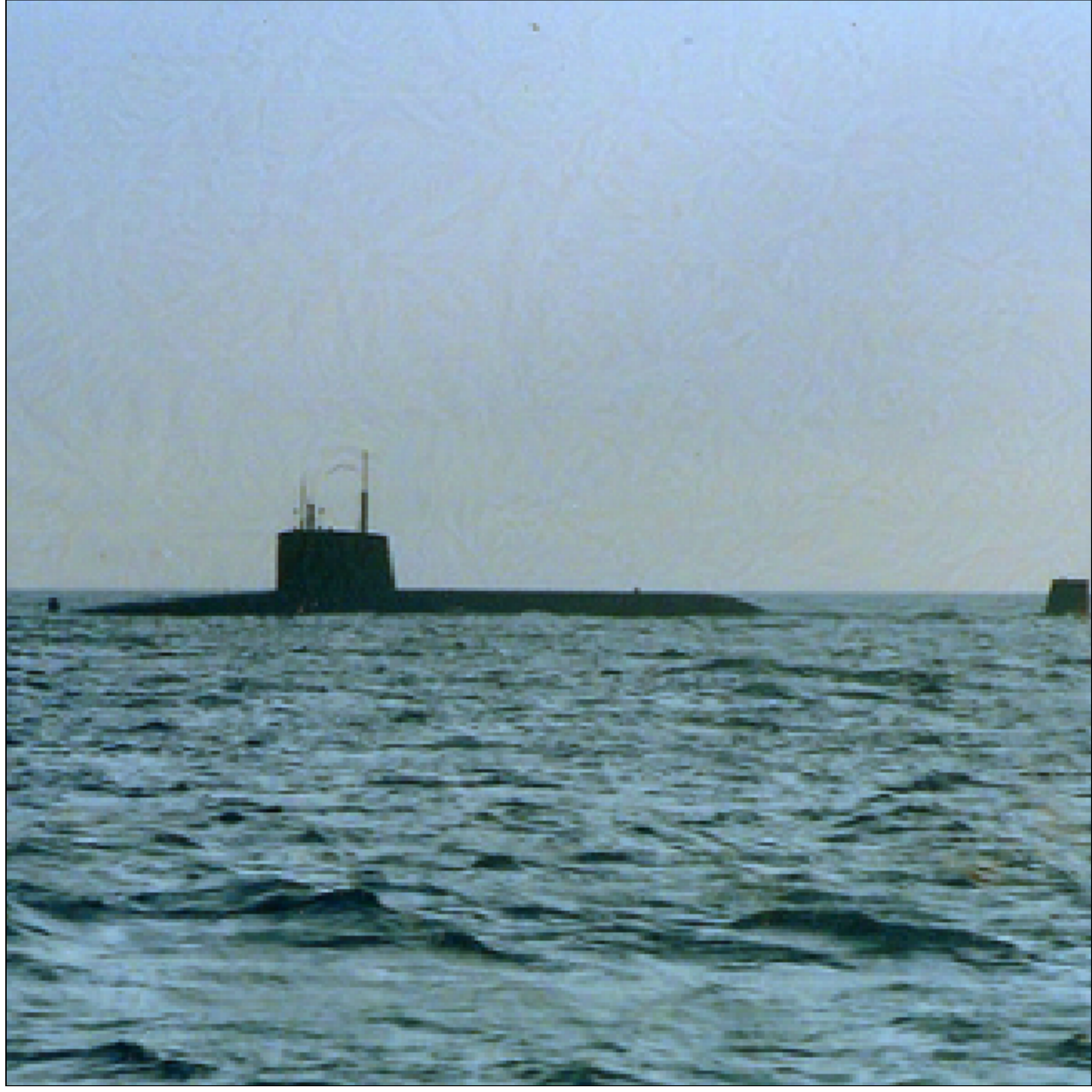}
\caption{FGSM}
  \end{subfigure}%
\\
\begin{subfigure}{.25\textwidth}
    \includegraphics[width=\textwidth]{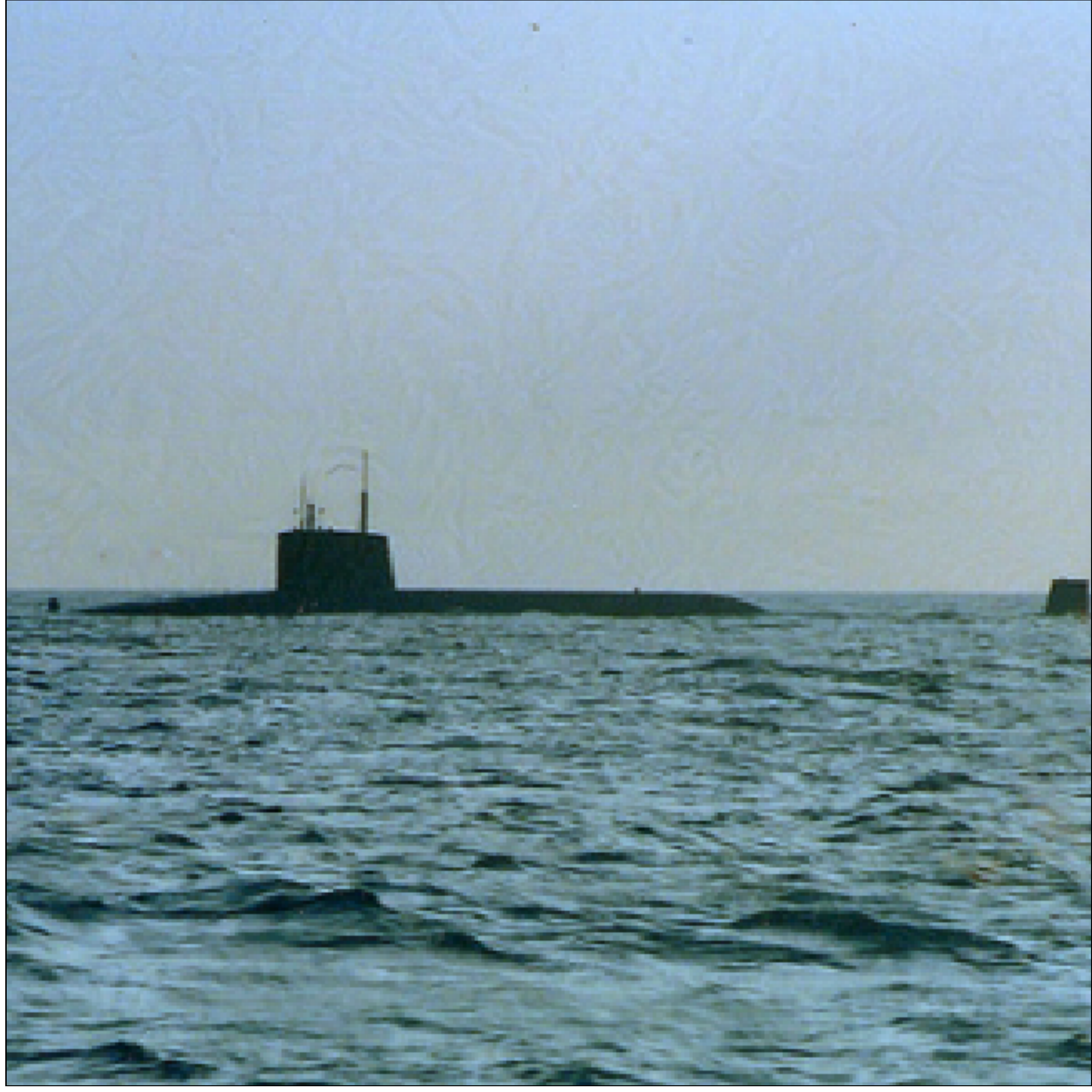}
\caption{Color-Aware}
  \end{subfigure}%
\begin{subfigure}{.25\textwidth}
    \includegraphics[width=\textwidth]{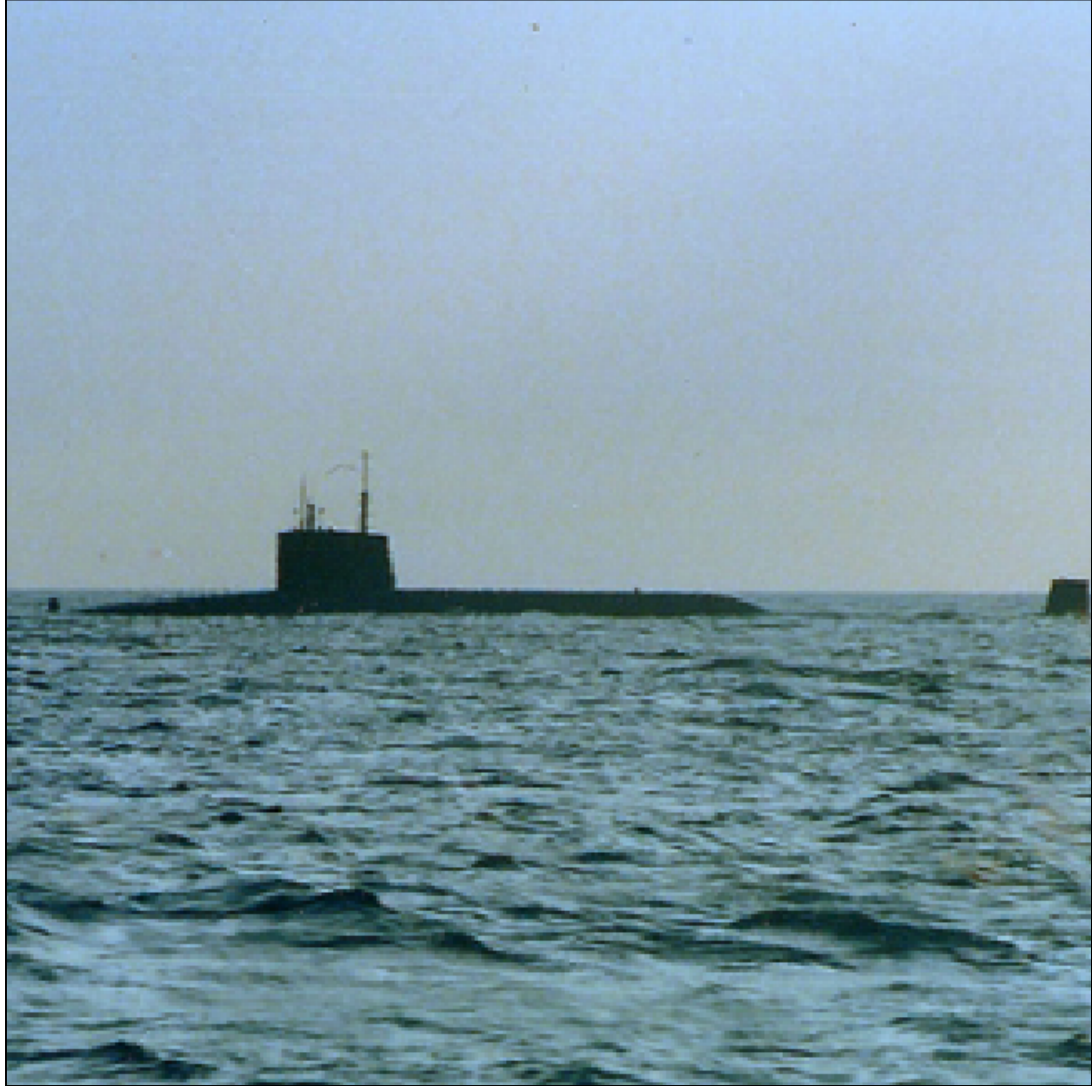}
\caption{Color-and-Edge-Aware}
  \end{subfigure}
\caption{Comparison of perturbation methods in an untargeted attack. For all methods the perturbed image was classified as a breakwater. Details in text and table \ref{subtable}.} 
\label{subfig}
\end{figure}

We also include a comparison of our Color-Aware and Color-and-Edge-Aware methods with FGSM and L-BFGS. In figure \ref{subfig} we construct untargeted perturbations on an image containing a submarine, where in this example and in the remainder of the paper we use the Sobel filter to construct the edge weights in Color-and-Edge-Aware. Close inspection reveals perturbation artefacts in the sky near the water's surface and around the submarine's broadcasting equipment for all methods except Color-and-Edge-Aware, with Color-Aware displaying less artefact than L-BFGS and FGSM. Table \ref{subtable} contains classification confidence and quantifications of the perturbation's size using various norms. The large $\ell_{2}$ norm of the Color-and-Edge-Aware perturbation relative to the other methods, combined with its small $\ell_{1}$ norm, suggests that there are fewer perturbed pixels but that the magnitude of these perturbations is larger. This aligns with the intuition used to create the Edge-Aware weighting. Little or no change occurs in the smooth sky region, while larger magnitude perturbations are placed in the region containing the water where it cannot be readily perceived.

\begin{table}
\begin{center}
\begin{tabular}{cccc}
Image & Prob. & $\|\delta\|_{1}$ & $\|\delta\|_{2}$\\ \hline
Original & $.997$ & $0$ & $0$\\
L-BFGS & $.442$ & $538.28$ & $1.89$\\
FGSM & $1.00$ & $1558.36$ & $3.82$\\
Color-Aware & $.980$ & $1594.55.69$ & $3.83$ \\
Color-\&-Edge-Aware & $.980$ & $987.29$ & $3.77$ 
\end{tabular}
\end{center}
\caption{Numerical comparison of the untargeted attacks producing the perturbed images in figure \ref{subfig}. For the original image the probability listed is for the submarine class. For all others it is the breakwater class, which was the class with the highest probability for all methods.}
\label{subtable}
\end{table}

\begin{figure}[h] 

\begin{center}   
\begin{subfigure}{.25\textwidth}     
\includegraphics[width=\textwidth]{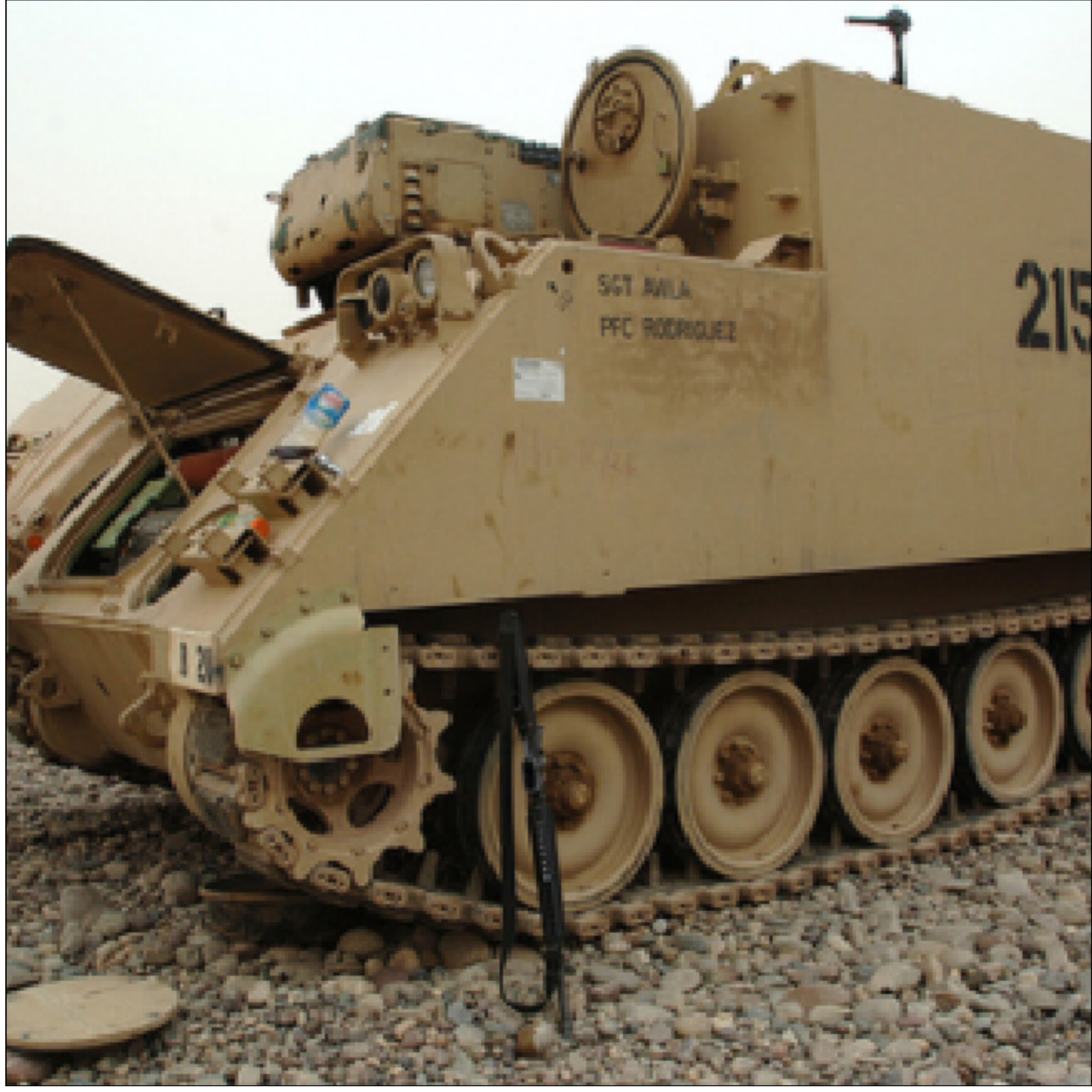} \caption{Original image}   
\end{subfigure}%
\end{center}   
\begin{subfigure}{.25\textwidth}     
\includegraphics[width=\textwidth]{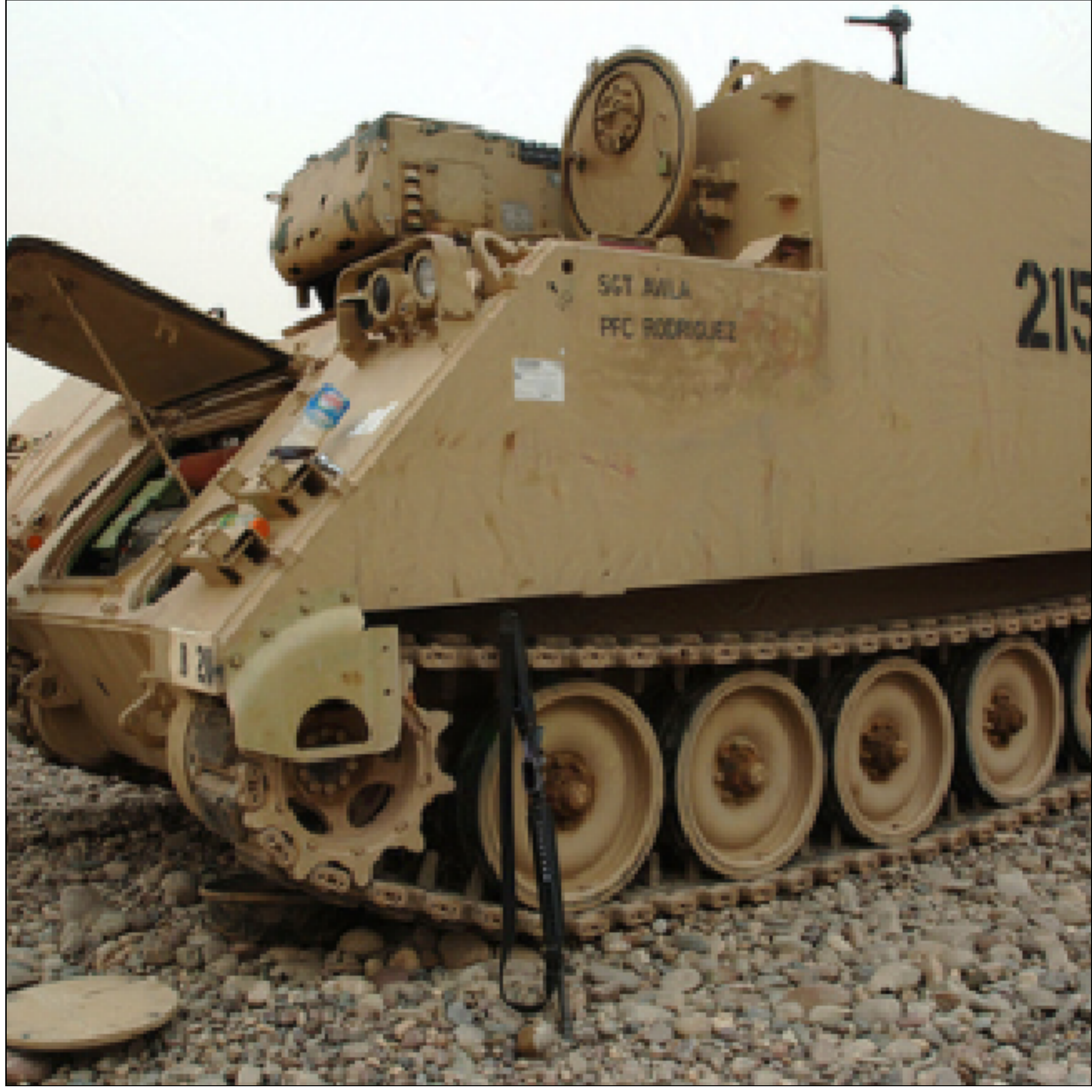} \caption{L-BFGS}  
\end{subfigure}%
\begin{subfigure}{.25\textwidth}     
\includegraphics[width=\textwidth]{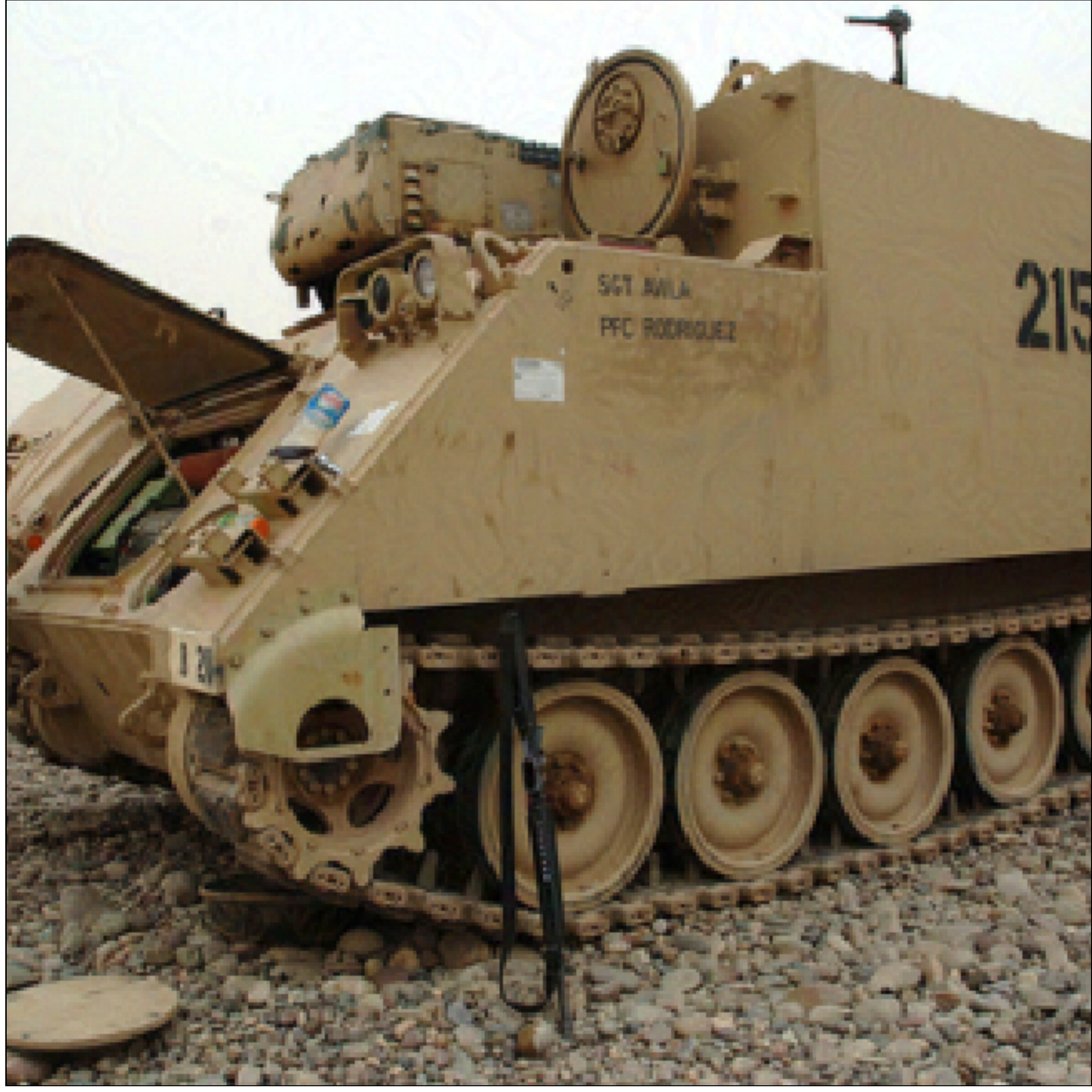} \caption{FGSM}   
\end{subfigure}%
\\ 
\begin{subfigure}{.25\textwidth}     
\includegraphics[width=\textwidth]{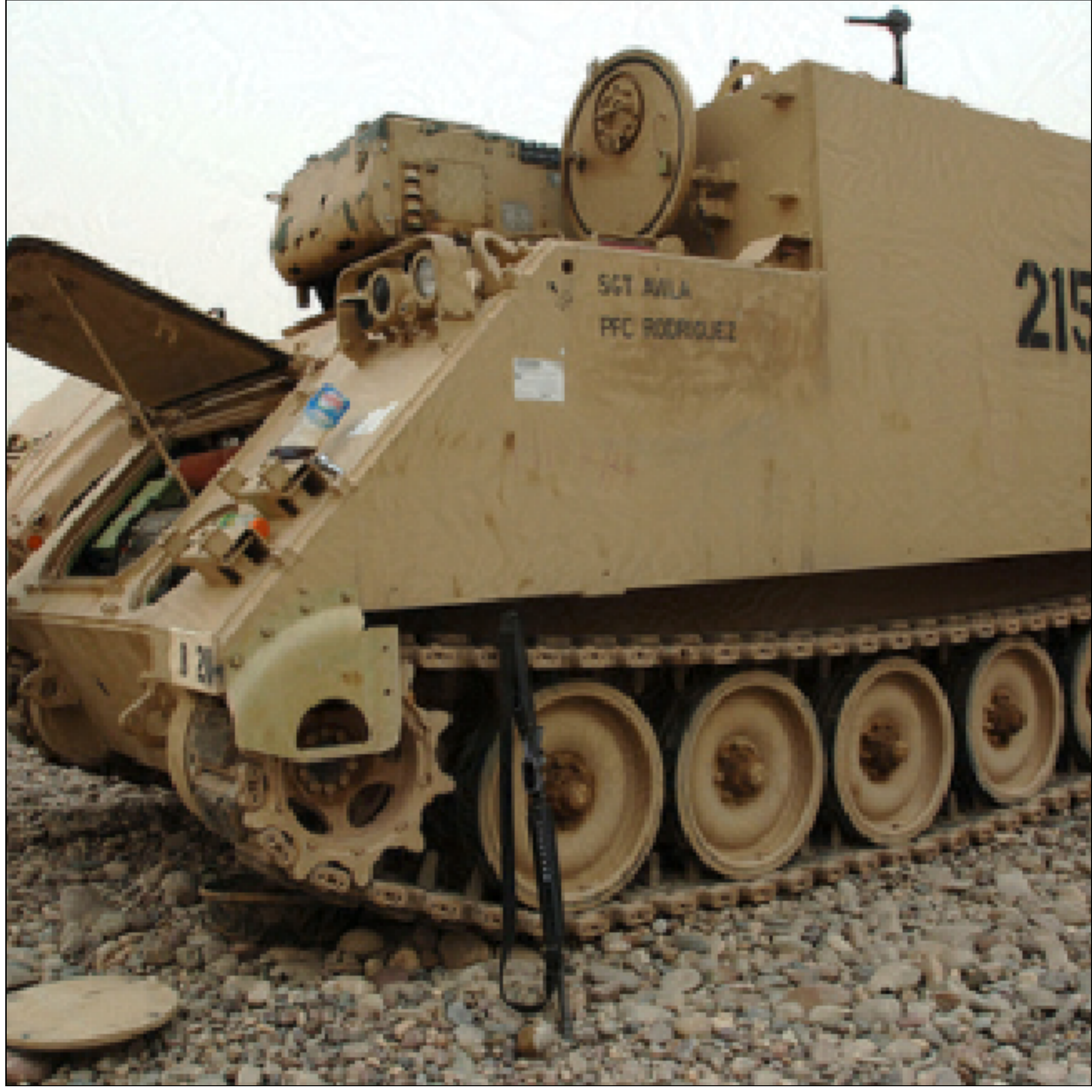} \caption{Color-Aware}   
\end{subfigure}%
\begin{subfigure}{.25\textwidth}     
\includegraphics[width=\textwidth]{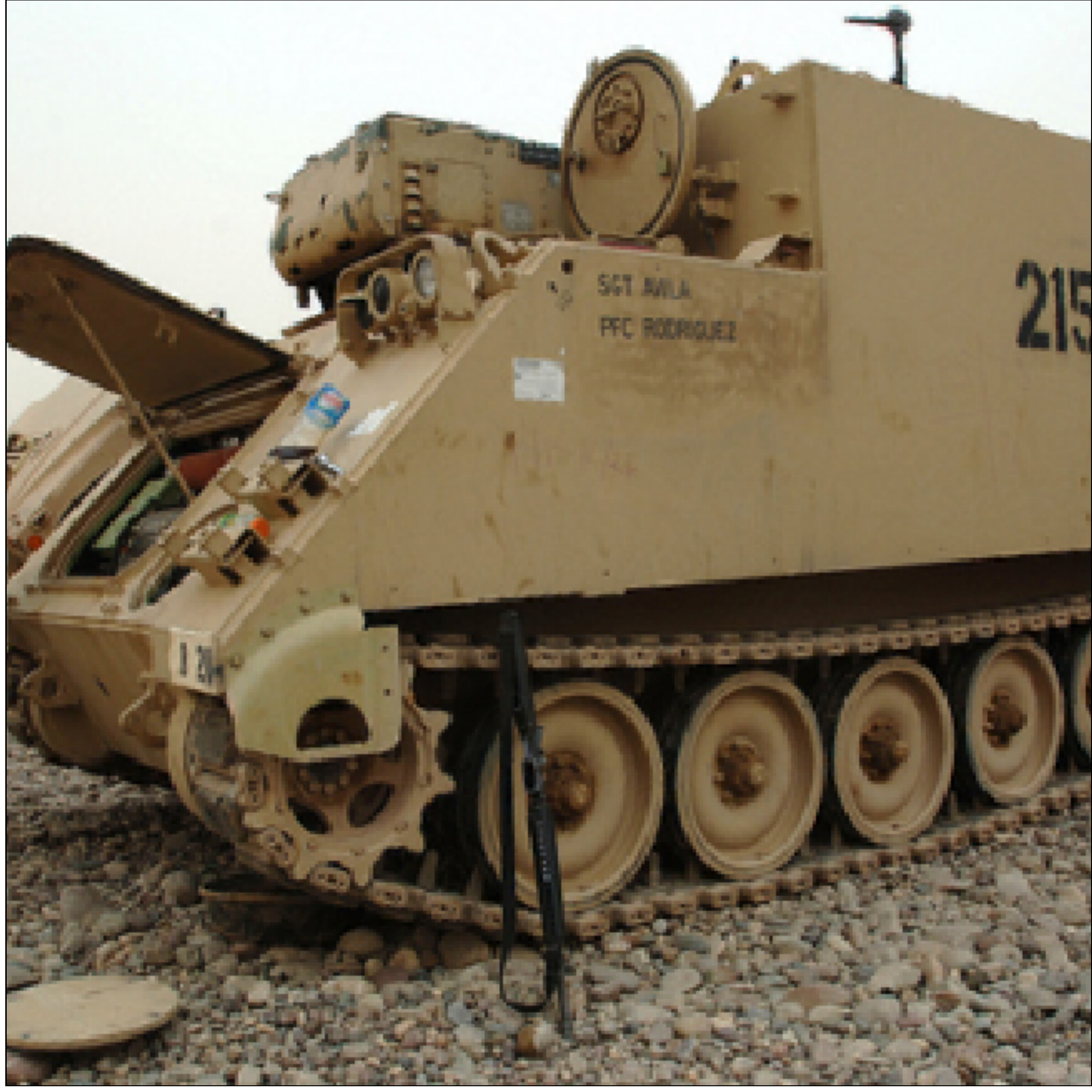} \caption{Color-and-Edge-Aware}   
\end{subfigure} \caption{Comparison of perturbation methods in a targeted attack, with a mobile home as the target label. Details in text and table \ref{tanktable}.} \label{tankfig} 
\end{figure}

Figure \ref{tankfig} compares perturbation methods in a targeted attack similar to the untargeted example in figure \ref{subfig}, where the source image contains a tank and the targeted label is a mobile home. All perturbation methods successfully induce the targeted misclassification. The performance of the methods is generally similar to the untargeted setting, in that all methods except Color-and-Edge-Aware have easily detectable perturbations. Similar to the submarine example, L-BFGS appears to place large magnitude perturbations in certain regions, in this example at the front of the tank, but less perturbation overall. Both Color-Aware and FGSM make notable texture changes on the side of the tank and in the sky above. Table \ref{tanktable} summarizes the performance of the methods. Though the Color-and-Edge-Aware perturbation is larger in $\ell_{2}$ norm than the others, suggesting the perturbation contains more extreme values, the location of these perturbations and its more accurate modeling of color perception makes it less discernible than the others from the perspective of a human observer.

\begin{table}[h]
\begin{center} 
\begin{tabular}{cccc} 
Image & Prob. & $\|\delta\|_{1}$ & $\|\delta\|_{2}$\\
\hline Original & $.852$ & $0$ & $0$\\
L-BFGS & $.804$ & $609.55$ & $1.79$\\
FGSM & $.995$ & $1621.22$ & $3.98$\\
 Color-Aware & $.980$ & $1616.69$ & $3.90$ \\
 Color and Edge-Aware & $.980$ & $1185.87$ & $4.45$ 
\end{tabular} 
\caption{Numerical comparison of the targeted attacks producing the perturbed images in figure \ref{tankfig}. For the original image the probability listed is for the tank class. For all others it is the mobile home class, which was the targeted label.} 
\label{tanktable} \end{center}
 \end{table} 

\begin{figure}[h]
\begin{center}
  \minipage{.25\textwidth}
    \includegraphics[width=\textwidth]{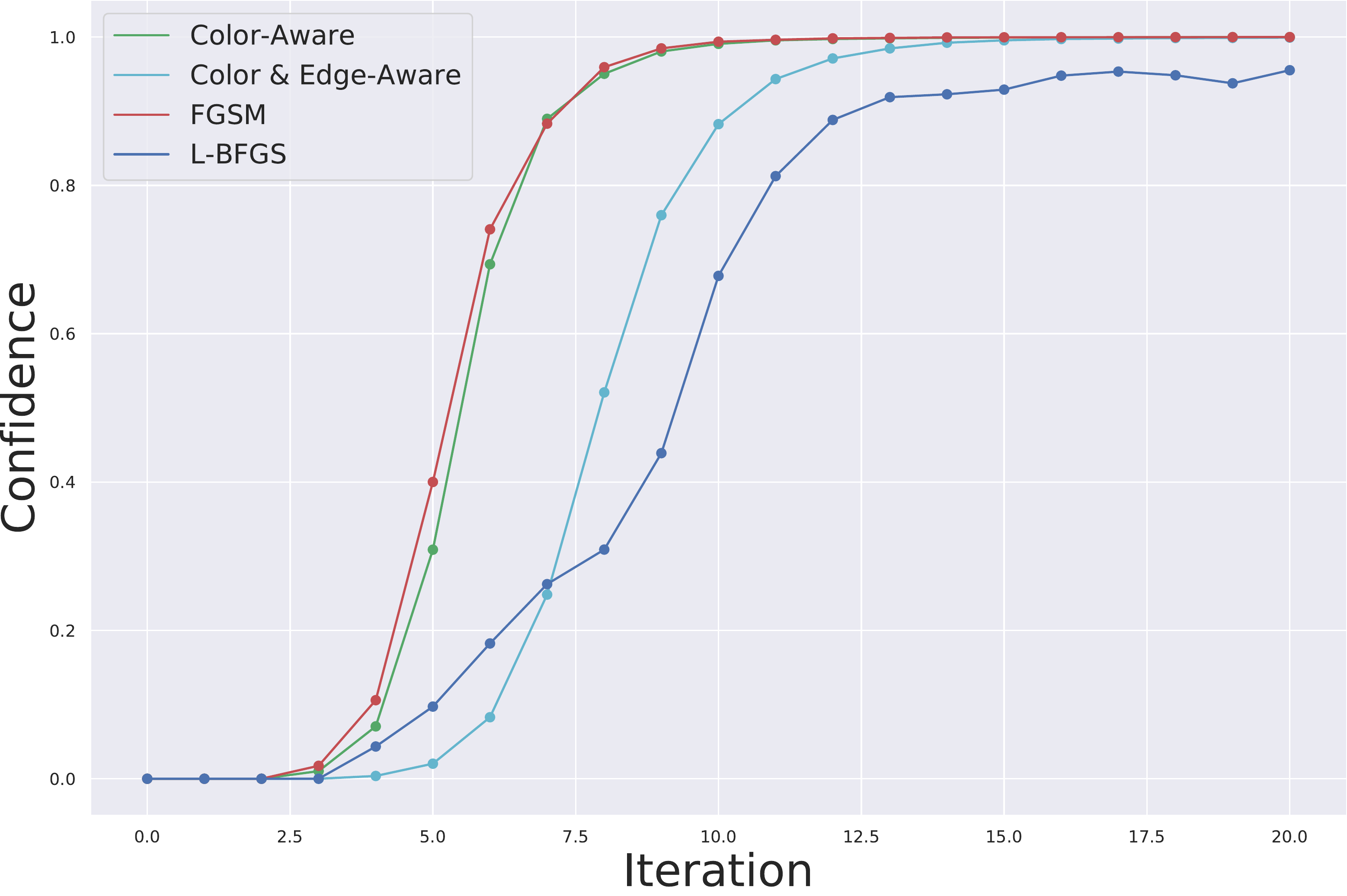}
  \endminipage%
\end{center}
\minipage{.25\textwidth}
    \includegraphics[width=\textwidth]{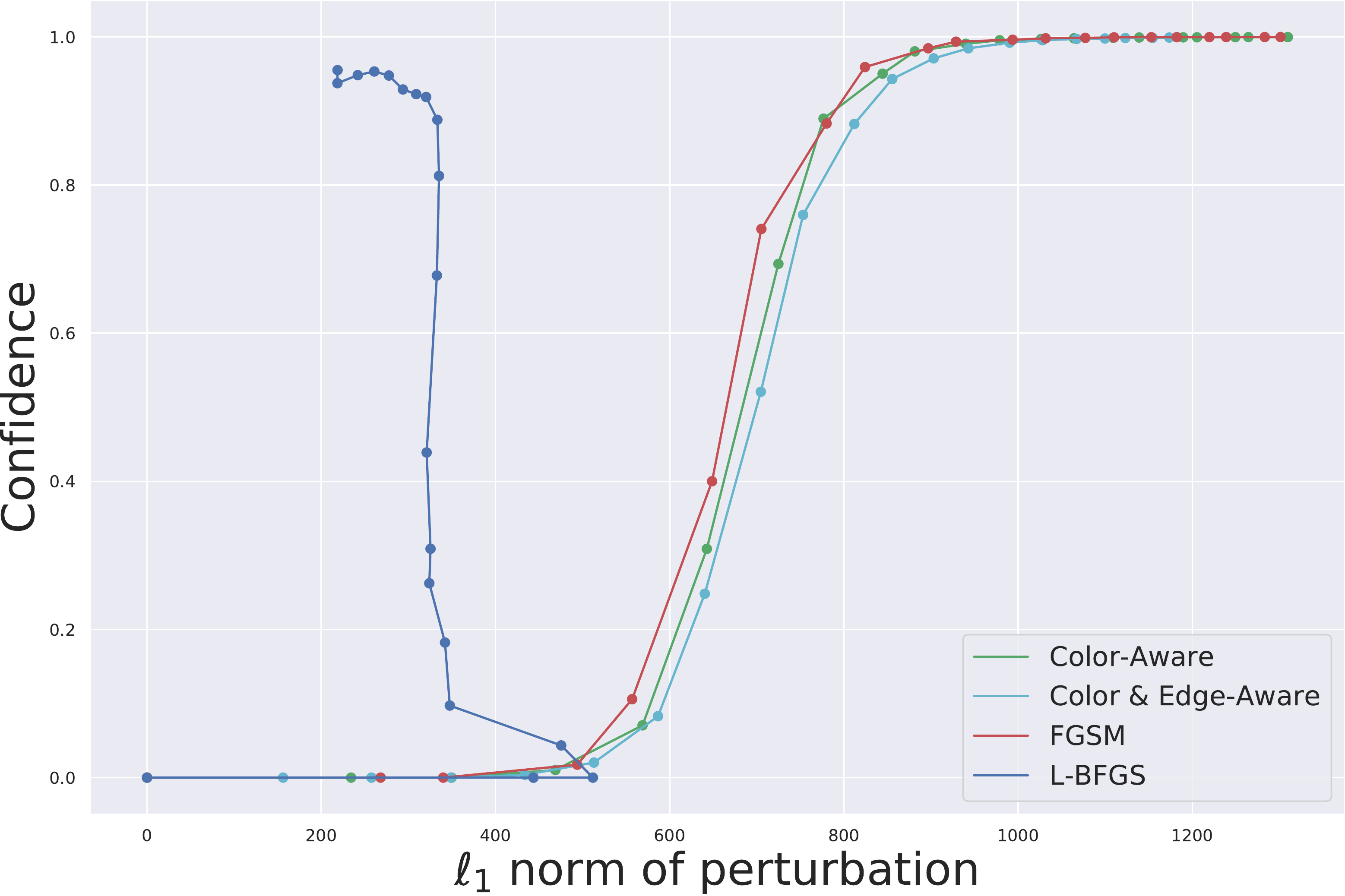}
  \endminipage
\minipage{.25\textwidth}
    \includegraphics[width=\textwidth]{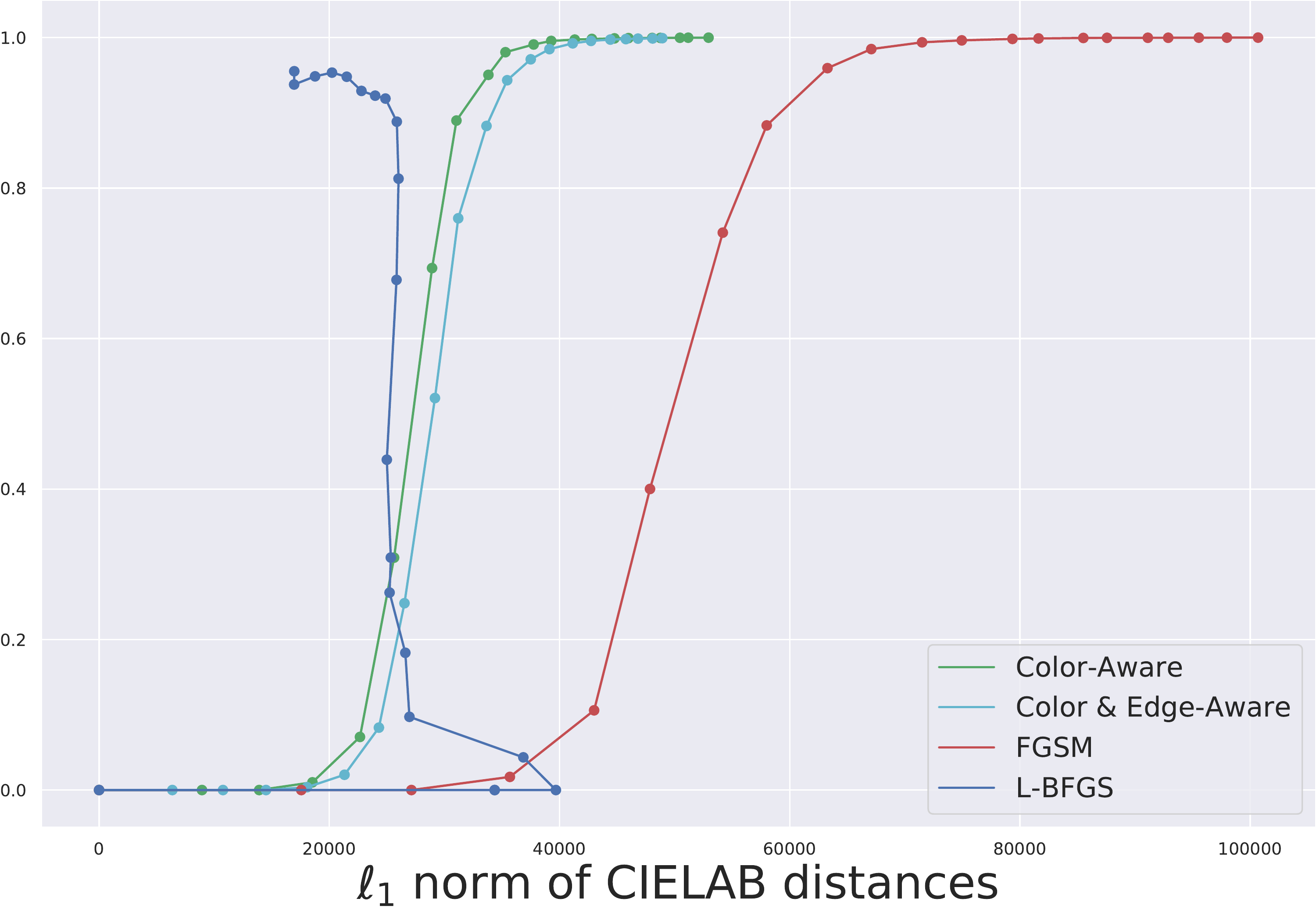}
  \endminipage\\
\minipage{.25\textwidth}
    \includegraphics[width=\textwidth]{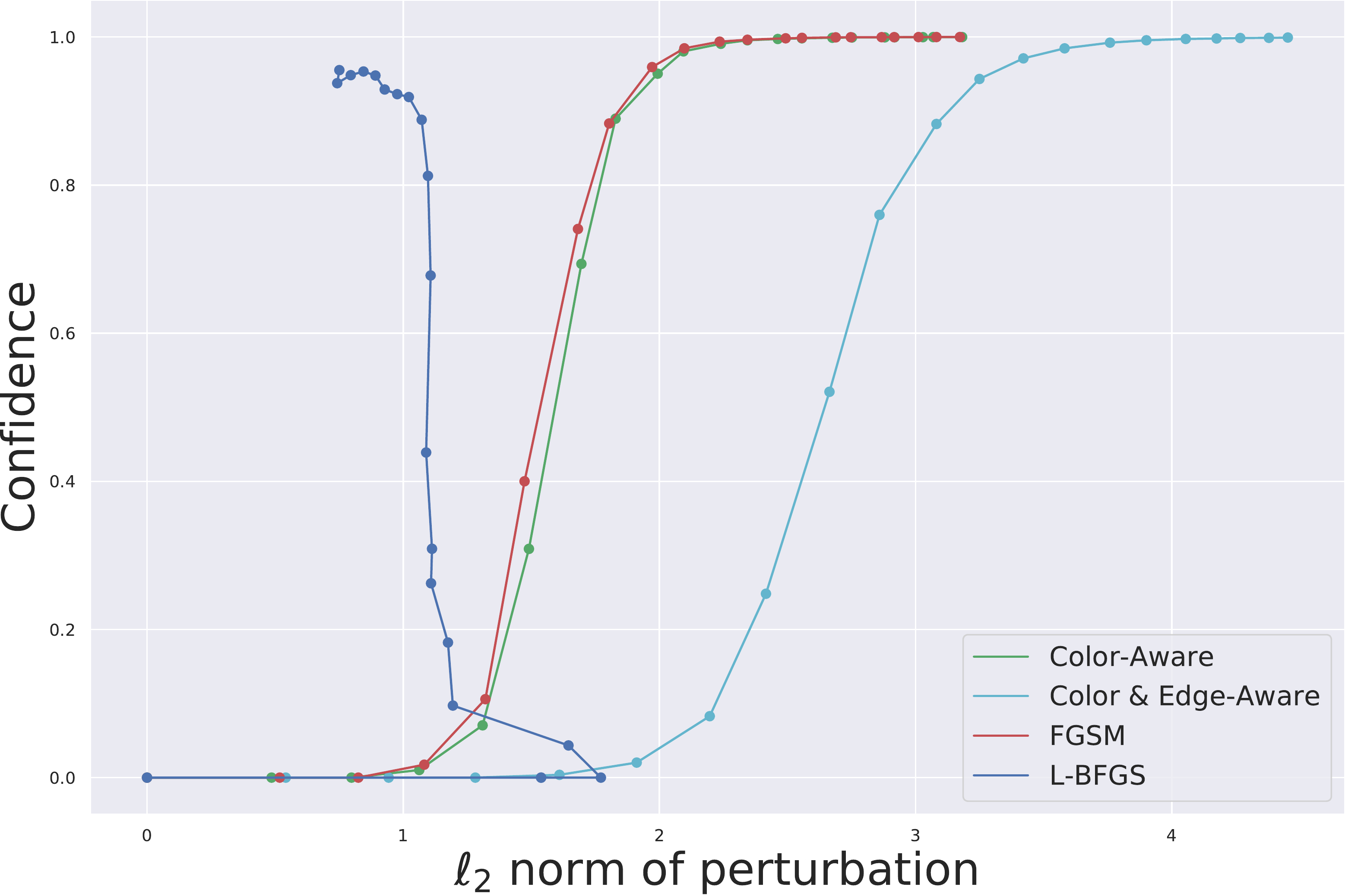}
  \endminipage
\minipage{.25\textwidth}
    \includegraphics[width=\textwidth]{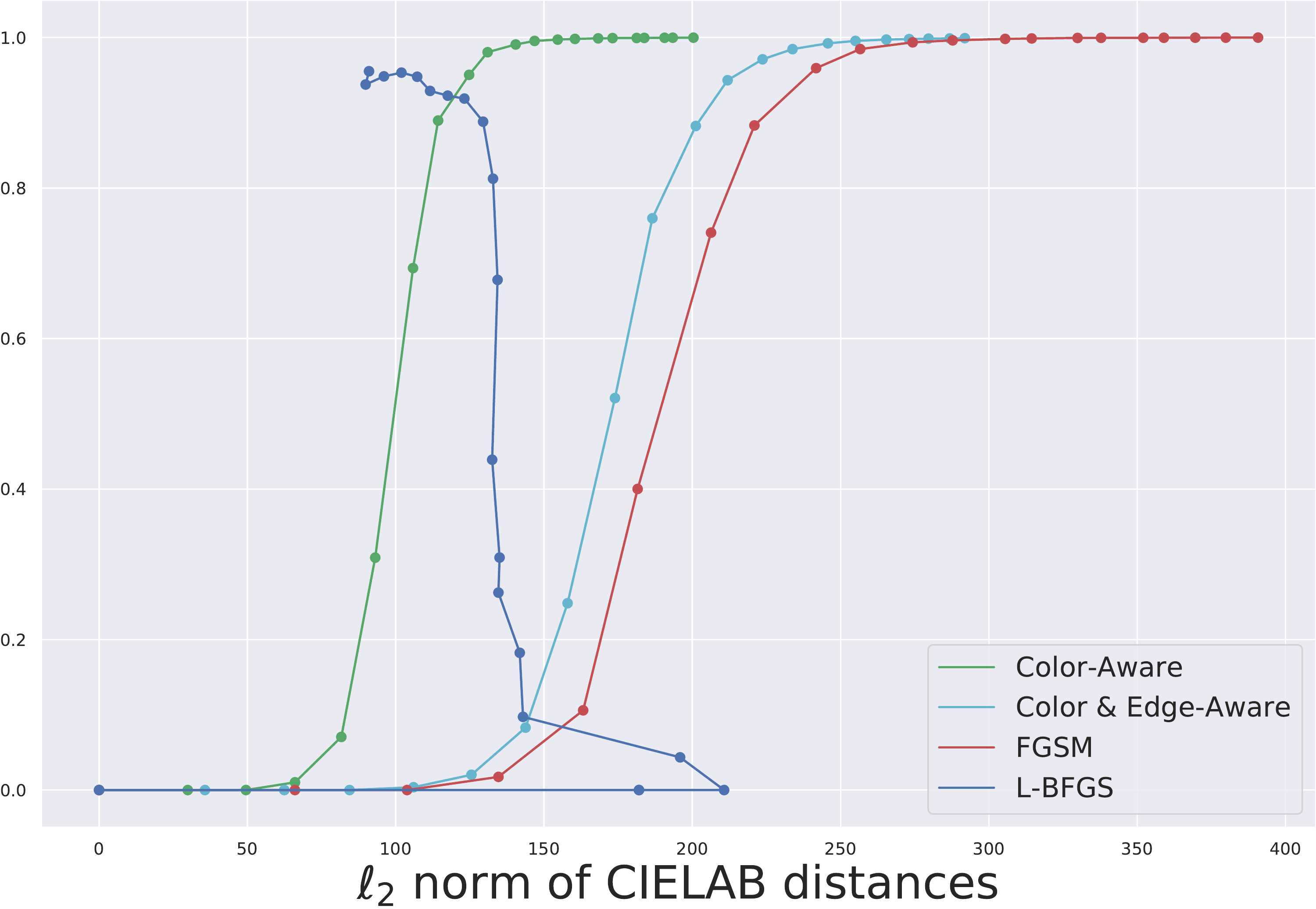}
  \endminipage\\
\minipage{.25\textwidth}
    \includegraphics[width=\textwidth]{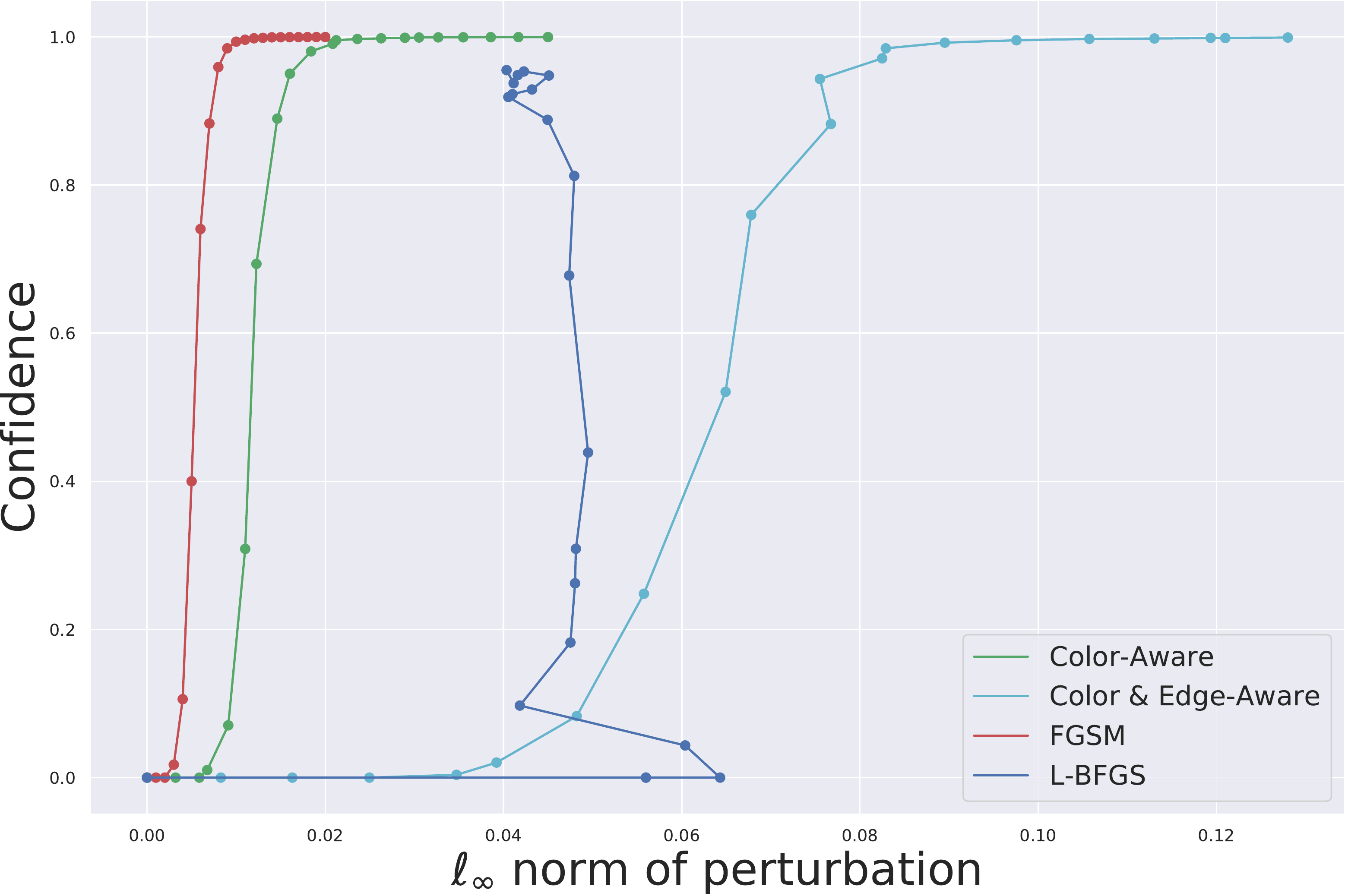}
  \endminipage
\minipage{.25\textwidth}
    \includegraphics[width=\textwidth]{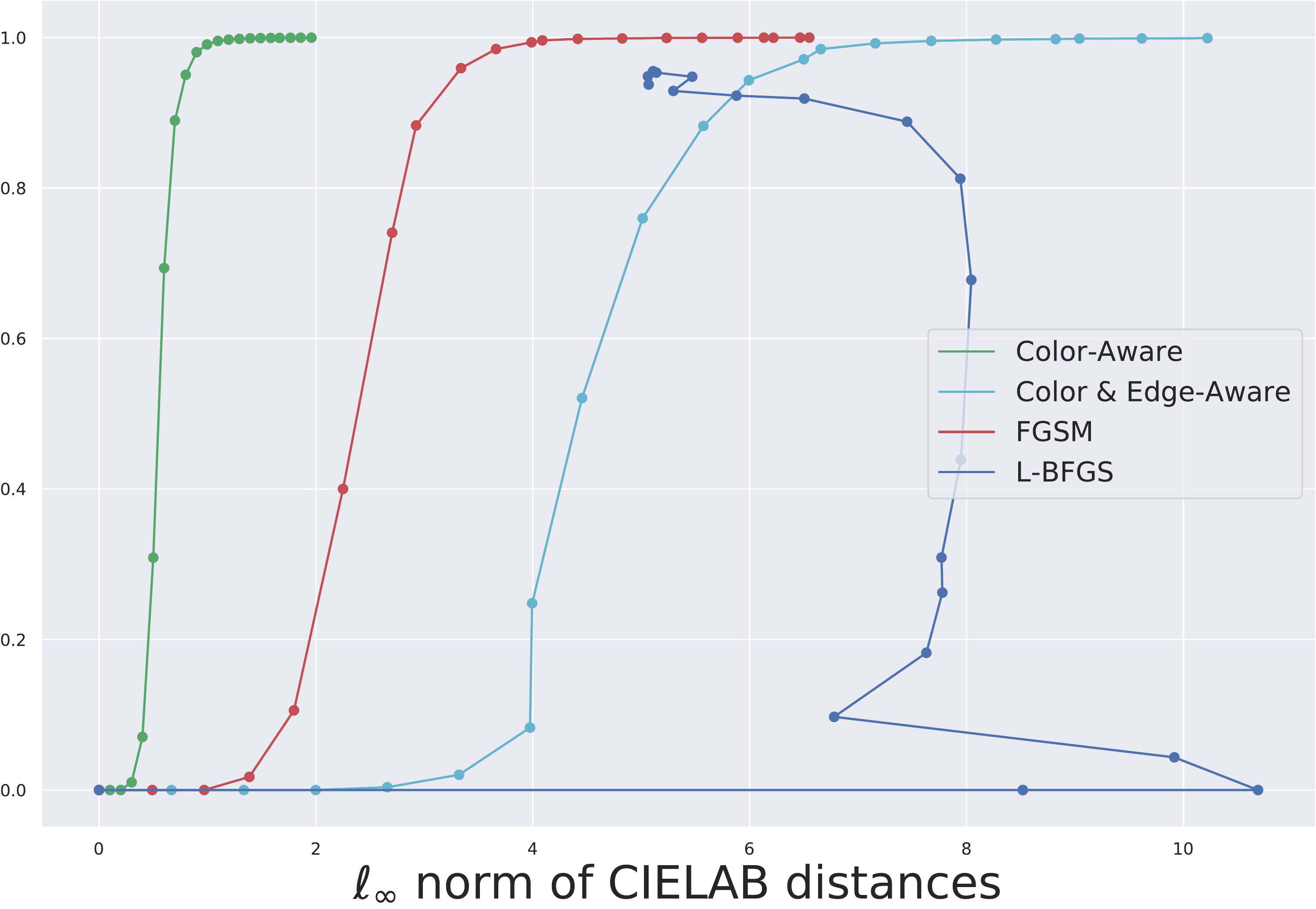}
  \endminipage\\

\caption{Confidence against iterations, $\ell_{1}$, $\ell_{2}$, and $\ell_{\infty}$ norms. The left column contains plots confidence against $\| x'^{\; \RGB} - x^{\RGB} \|_{p}$  and the right column plots confidence against $\left\| \| x'^{\; \LAB} - x^{\LAB}\|_{2,c} \right\|_{p}$, for $p \in \{1, 2, \infty\}$. Details in text.}
\label{conffigs}
\end{figure}

As a test of the efficiency of our contributions, we next compare how the misclassification confidence changes as a function of the number of iterations and magnitude of the perturbation permitted. We use the source image of a submarine in figure \ref{subfig} to perform an untargeted attack. Our results are given in figure \ref{conffigs}, where the curves in each subfigure are parametrized by iteration number. These plots show that our contributions are competitive with L-BFGS and FGSM when measured by the $\ell_{p}$ norms in an RGB representation, and that they \emph{outperform} the competition when quantified using the CIELAB color distance.

From the perspective of per-iteration confidence, figure \ref{conffigs} provides evidence that FGSM and Color-Aware perturbations behave similarly with respect to misclassification confidence as a function of iterations. Color-and-Edge-Aware performs poorly in metrics which are sensitive to outliers ($\ell_{2}$ and $\ell_{\infty}$) because of its tendency to place larger perturbations in regions they are not easily detected. L-BFGS also performs poorly in these metrics, suggesting that it places large perturbations in isolated regions. That fact that L-BFGS does not place these perturbations in regions where they are difficult to discern suggests that the perturbation artefacts may be easily detectable. When quantified using the $\ell_{1}$ norm of the CIELAB distances, Color and Edge-Aware perturbations outperform FGSM and are competitive with the more complex L-BFGS method.

We next demonstrate that Color-Aware and Color-and-Edge-Aware perturbations are effective at inducing misclassification, while also making less perceptible change to the image. For our first experiment, we choose 100 images from the 2012 ILSVRC validation set which the Inception v3 network classifies correctly. The $\alpha$ value is chosen uniquely from a set of candidates for each perturbation method but fixed for all images. Our selected $\alpha$ corresponds to the value that misclassified the highest proportion of the first 10 images. We perform both targeted and untargeted experiments, where for all images the target class was a coffee mug. The step length/penalty parameter was chosen in the untargeted experiment and was the same for the targeted experiments. Table \ref{misclass_tab} summarizes our results.

\begin{table}[H]
\begin{tabular}{c|cccc}
& L-BFGS & FGSM & C-Aware & C \& E-Aware\\\hline
Untargeted & $92\%$ & $100\%$ & $100\%$ & $91\%$\\
Targeted & $100\%$ & $100\%$ & $100\%$ & $91\%$
\end{tabular}
\caption{Misclassification percentages for 100 images with 5 iterations for untargeted and 10 iterations for targeted. Full details in text.}
\label{misclass_tab}
\end{table}
\begin{figure}
\begin{center}
\includegraphics[width=.5\textwidth]{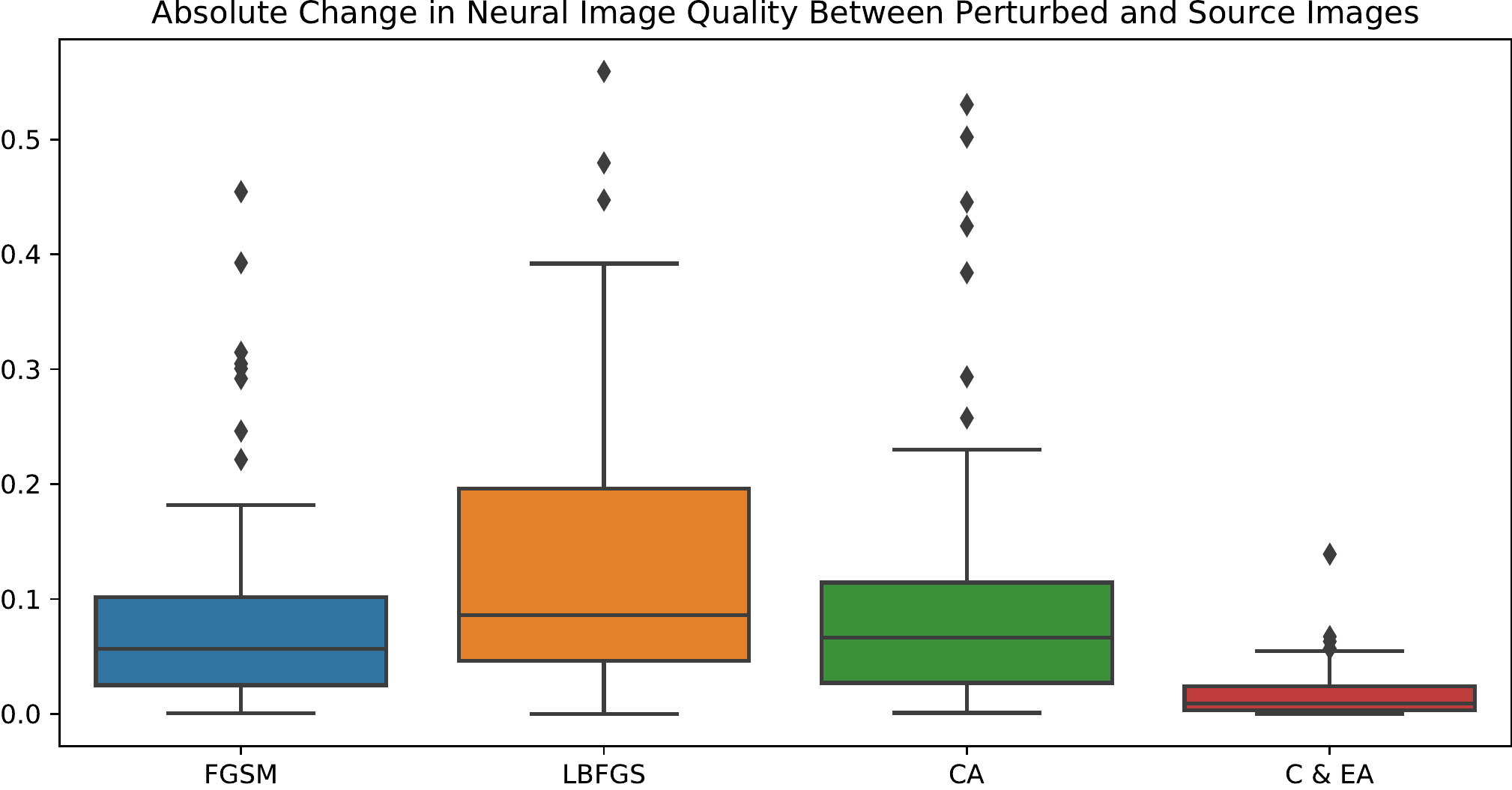}
\end{center}
\caption{Change in NIMA quality for the 100 perturbed images in the untargeted example from table \ref{misclass_tab}. See text for details.}
\label{fig:nima}
\end{figure}

The results indicate that perturbations which are Color-Aware and/or Edge-Aware reliably induce misclassification. Moreover, all methods considered consistently cause misclassification given proper choice of $\alpha$ and sufficient number of iterations. Unsurprisingly, we see that achieving targeted misclassification is more difficult and requires more iterations to do so reliably. Our Color-and-Edge-Aware method achieves $91\%$ misclassification rate on both the untargeted and targeted tasks whereas FGSM and Color-Aware achieves $100\%$ misclassification, providing evidence that restricting perturbations to smoothly-textured regions has only a small impact on the ability of the method to induce misclassification. Only $3\%$ of images were unable to be misclassified by \emph{both} untargeted and targeted Color-and-Edge-Aware, which suggests that ability to induce misclassification using perturbations which are restricted to certain regions depends on the region and the target class. 

Having effectively induced misclassification, we next discuss the perceptibility of the perturbations in table \ref{misclass_tab}. As a substitute for the large scale human studies, we compare the perceived difference between the perturbed and original images using Neural Image Assessment (NIMA) \cite{nima}, which has been shown to reliably predict human opinion scores of an image's quality, which we expect to degrade with the introduction of perturbation artifacts. Figure \ref{fig:nima} shows the perceived difference, quantified as the absolute change in NIMA quality, for the 100 images in the untargeted example from table \ref{misclass_tab}. We see that, among all the perturbation methods considered, our Color-and-Edge-Aware method produces perturbed images which has minimizes the perceived difference from the original.

%
%
%

Next we compare the computational burden of the methods considered, as assessed by their running times. Color-Aware and Color-and-Edge-Aware perturbations have the advantage of properly accounting for human perception of color and texture, but we hope to do so as simply as possible. The gold standard for generating simple image perturbations is the fast gradient sign method, because it only requires the elementwise sign of the gradient. Compared to FGSM, Color-Aware and Color-and-Edge-Aware impose additional structure to model human perception. First, the computation of the gradient requires composition with the conversion function $F_{\LAB \to \RGB}$. Second, instead of the sign of the gradient Color-Aware methods normalize the gradient in the $\ell_{2}$ norm across the color dimension. Lastly, making a perturbation method Edge-Aware requires an additional application of an edge filter. Our experiments show that Color-Aware and Color-and-Edge-Aware perturbations are only marginally less efficient to construct than FGSM perturbations.

Because our emphasis is on generating effective image perturbations with minimal computational resources, we consider both CPU and GPU implementations. Table \ref{tab:runtimes} gives the running times for each method applied to 100 ILSVRC images. Our experiments were carried out in a Linux operating system, with an 8-core Intel i9-9980 CPU at 2.4 GHz, 64 GB of memory, and a GeForce GTX 1650 GPU.

\begin{table}[H]
\begin{center}
\begin{tabular}{ccc }
Method & CPU time (s) & GPU time (s) \\\hline
L-BFGS & $3.56 \pm 0.55$ & $0.52 \pm 0.04$ \\
FGSM & $1.71 \pm 0.05$ & $0.29 \pm 0.04$ \\
Color-Aware & $1.81 \pm 0.12$ & $0.35 \pm 0.04$\\
Color-\&-Edge-Aware & $1.83 \pm 0.03$ & $0.36 \pm 0.06$
\end{tabular}
\caption{Mean and standard deviation of the run times for each method applied to 100 images. Each method was run for 10 iterations.}
\label{tab:runtimes}
\end{center}
\end{table}

Table \ref{tab:runtimes} demonstrates that the time required to generate Color-Aware and Color-and-Edge-Aware perturbations is similar to FGSM and less than L-BFGS. Interestingly, we note that the computational burden associated with composing the model with the conversion function $F_{\LAB \to RGB}$, evident in the difference between FGSM and Color-Aware, results in a smaller relative increase in the CPU implementation ($5.8\%$) than the GPU implementation ($20.7\%$). In both implementations, the additional time required to make the method edge-aware is minor, between $0.01$ and $0.02$ seconds.

\subsection{Perturbing DeepFakes}

Our next set of numerical experiments focuses on adversarially perturbing artificially generated images to reduce their risk of detection. We focus our attention on \emph{DeepFakes}, in which an individual in an image is replaced by another person using an autoencoder-based face swap. DeepFake production continues to progress in sophistication and ease of use, resulting in the proliferation of increasingly convincing forgeries. Because of their potential for malicious use, developing methods to distinguish DeepFakes from authentic images is an active area of research.

One promising tool for DeepFake detection is proposed in \cite{CNN_easy}, where the authors show that DeepFakes and many other types of synthetic images can be reliably detected using a ResNet-50 architecture configured for binary classification and trained on images from ProGAN, an unconditional GAN-based generator. We show that applying Color-and-Edge-Aware perturbations to DeepFakes results in images which the classifier cannot distinguish from authentic images. Our contributions are appropriate for this application because we perturb images of faces, where human perception is especially sensitive to manipulations.


\begin{figure}[h]
        \setlength{\tabcolsep}{1.5pt}
        \begin{tabular}{cc}
               \scriptsize{Original Image} & \scriptsize{DeepFake}\\
               \subcaptionbox*{\scriptsize{}\label{a}}{\includegraphics[width=.2445\textwidth]{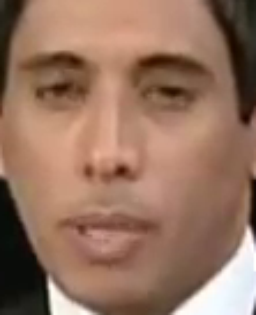}} &  \includegraphics[width=.25\textwidth]{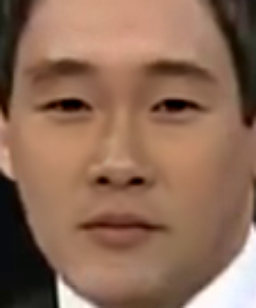}
\vspace{-.6cm}
                 \\
 \scriptsize{$P(\text{DeepFake}) < 0.0001$} & \scriptsize{$P(\text{DeepFake}) > .9999$} \\
\scriptsize{Perturbation} & \scriptsize{Perturbed Result} \\
            \subcaptionbox*{\scriptsize{}\label{b}}{\includegraphics[width=.25\textwidth]{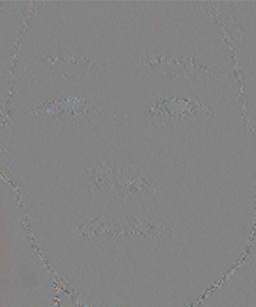}} &    \includegraphics[width=.25\textwidth]{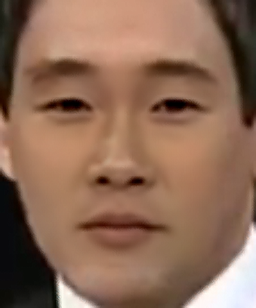}
\vspace{-.6cm}
\\
               & \scriptsize{$P(\text{DeepFake}) < 0.0001$} \\
        \end{tabular}
\caption{Application of Color-and-Edge-Aware perturbation to DeepFake Detection. Despite correctly identifying the labels of the original manipulated image, the DeepFake detector fails to identify the manipulated image after applying a perturbation which avoids placing artefacts in regions of the face where it would be easily detected.}
\label{deepfakefig}
\end{figure}

We consider the same set of DeepFakes and authentic images as \cite{CNN_easy}, 5405 faces extracted from video frames in the FaceForensics++ DeepFake data set \cite{FF++}. For these images, the proposed DeepFake detector achieves an impressive Average Precision (AP) score of $.982$. By adding Color-and-Edge-Aware perturbations to these images, we are effectively creating a post-processing filter which allows these forgeries to escape detection. Figure \ref{deepfakefig} shows one such perturbation, where the classifier correctly classifies the original and manipulated images, but the perturbed DeepFake is classified as authentic. Note that the perturbation respects the structure of the human face by leaving its inner region largely unaltered.



As a further demonstration of the utility of Color-and-Edge-Aware perturbations in this application, we construct a perturbation for each of the authentic and DeepFake images in the data set\footnote{The $\alpha$ value and iteration chosen are selected by hand tuning over a small list of possible candidates. We use $\alpha=10$ and $15$ iterations for all $5405$ perturbations, including Figure \ref{deepfakefig}.}. Despite only making minimal changes to the depicted faces, table \ref{tab:APs} shows that these perturbations completely eliminate the utility of the DeepFake detector, resulting in predictions which are worse than choosing the predicted label uniformly at random.

\begin{table}
\begin{center}
\begin{tabular}{c|cc}
& Unperturbed & Perturbed\\ \hline
Average Precision & $.982$ & $.37$
\end{tabular}
\caption{Average precision of the ResNet model \cite{CNN_easy} for DeepFake detection before and after applying Color-and-Edge-Aware perturbations.}
\label{tab:APs}
\end{center}
\end{table}

\section{Conclusion}

We have presented two new methods for creating adversarial image perturbations which are less discernible by a human observer. The first, our \emph{Color-Aware} perturbation method, accounts for human perception of color by performing the perturbation directly in CIELAB space, where a simple constraint guarantees the perceived color change to the perturbed image is small. Our second contribution is our Edge-Aware method, which uses a texture filter to restrict perturbations to regions where a human observer is less likely to detect them. Color-Aware and Edge-Aware methodology can be combined to generate \emph{Color-and-Edge-Aware} perturbations, which address both issues simultaneously. We find that our contributions reliably induce misclassification, require similar computation time as the most efficient techniques for generating adversarial perturbations, and are more difficult to detect than methods of similar complexity, providing evidence that Color and Edge-Aware perturbations are a simple yet effective way to generate perturbations which properly account for human perception.

\section*{Acknowledgments}

All authors acknowledge support from 
the Office of Naval Research's Science of Autonomy Program, award number N0001420WX01523.

{\small
\bibliographystyle{template/ieee_fullname}
\bibliography{works_cited}
}

\end{document}